\documentclass[journal,twoside,web]{ieeecolor}
\usepackage{tmi}

\usepackage{cite}
\usepackage{amsmath,amssymb,amsfonts}
\usepackage{algorithmic}
\usepackage{graphicx}
\usepackage{textcomp}

\usepackage{diagbox}

\usepackage{amssymb}

\def\etal{~\textit{et.al.}}

\newcommand{\vect}[1]{\boldsymbol{#1}}

\usepackage{bm}
\usepackage{booktabs}

\usepackage{arydshln}

\def\BibTeX{{\rm B\kern-.05em{\sc i\kern-.025em b}\kern-.08em
    T\kern-.1667em\lower.7ex\hbox{E}\kern-.125emX}}
\begin{document}

% for numbers and formatting
% \linenumbers
% \linenumbersep 2pt\relax 

% \title{IFSS-Net: Interactive Few-Shot Siamese Network for Faster Muscles Segmentation and Propagation in 3-D Freehand Ultrasound}

\title{IFSS-Net: Interactive Few-Shot Siamese Network for Faster Muscle Segmentation and Propagation in Volumetric Ultrasound}

\author{Dawood Al Chanti, Vanessa Gonzalez Duque, Marion Crouzier, Antoine Nordez, Lilian Lacourpaille, and Diana Mateus

\thanks{This work has been supported in part by the European Regional Development Fund the Pays de la Loire region on the Connect Talent scheme (MILCOM Project) and Nantes Métropole.% (Convention 2017-10470).
}
\thanks{Dawood Al Chanti, Vanessa Gonzalez Duque and Diana Mateus are with École Centrale de Nantes, Laboratoire des Sciences du Numérique de Nantes LS2N, UMR CNRS 6004 Nantes, France (e-mail:dawood.alchanti@ls2n.fr).}
\thanks{Marion Crouzier, Antoine Nordez, and Lilian Lacourpaille are with Université de Nantes, Laboratoire ``Movement - Interactions - Performance", MIP, EA 4334, F-44000 Nantes, France.}
}

\maketitle

\begin{abstract}
%214 WORDS
%214 WORDS
We present an accurate, fast and efficient method for segmentation and muscle mask propagation in 3D freehand ultrasound data, towards accurate volume quantification. 
A deep Siamese 3D Encoder-Decoder network that captures the evolution of the muscle appearance and shape for contiguous slices is deployed. We uses it to propagate a reference mask annotated by a clinical expert.  To handle longer changes of the muscle shape over the entire volume and to provide an accurate propagation, we devise a Bidirectional Long Short Term Memory module.  Also, to train our model with a minimal amount of training samples, we propose a strategy combining learning from few annotated 2D ultrasound slices with sequential pseudo-labeling of the unannotated slices. 
We introduce a decremental update of the objective function to guide the model convergence in the absence of large amounts of annotated data. After training with a small number of volumes, the decremental update transitions from a weakly-supervised training to a few-shot setting. 
Finally, to handle the class-imbalance between foreground and background muscle pixels, we propose a parametric Tversky loss function that learns to adaptively penalize false positives and false negatives. We validate our approach for the segmentation, label propagation, and volume computation of the three low-limb muscles on a dataset of 61600 images from 44 subjects. We achieve a Dice score coefficient of over $95~\%$ and a volumetric error \textcolor{black}{of} $1.6035 \pm 0.587~\%$.
\end{abstract}


\section{Introduction}
\label{sec:introduction}

% DA:
% -------------------------
% Blue Color is new comments

% Start with the problem description from biomedical point of view (Application)
\IEEEPARstart{Q}{uantification} of muscle volume is a useful biomarker for degenerative neuromuscular disease progression or sports performance\cite{pichiecchio2018muscle}.
%useful in understanding the evolution of certain diseases but also the effect of sportive training. For instance, severity and progression of degenerative neuromuscular diseases, such as Duchenne Muscular Dystrophy (DMD), can be captured by evaluating the fat infiltration of muscle tissue. Moreover, in sport, it is of interest to follow muscle volume change~\cite{pichiecchio2018muscle} to optimize the expected results in gaining strength and muscle mass due to repeated training sessions. 
%
% Treatment follow-up, to evaluate and reveal the underlying anatomical and morphological changes in patient's muscle volume, often requires the segmentation of a 3D images of the lower limb muscles. This work focuses on computing the volume fraction of the low limb muscles, mainly: the Gastrocnemius Medialis (GM), the Lateralis (GL), and the Soleus (SOL), for assessing disease progression of DMD and to aid volume computation in sport.
%
%Measuring the underlying anatomical and morphological changes in patient's muscle volume, often requires the segmentation of a 3D images. This work focuses on computing the volume fraction of the low limb muscles, mainly: the Gastrocnemius Medialis (GM), the Lateralis (GL), and the Soleus (SOL), with the objective of later assessing disease progression in DMD and assisting volume measurement during sportive training.
Measuring muscle volume often requires the segmentation of 3D images. While  Magnetic Resonance (MR) is the modality of preference for imaging muscles, 3D Ultrasound (US) offers a real-time, inexpensive, and portable alternative.
The motivation of our work is to assist the segmentation and volume computation of the low limb muscles from 3D freehand ultrasound volumes.
However, the methods here developed are general and may be of interest for other clinical applications requiring the segmentation of organs in 3D ultrasound images~\cite{gomariz2019siamese,dunnhofer2020siam}, \textcolor{black}{as well for other modalities such MRI \cite{Novikov2018tmi} or CT volumes \cite{dou20173d}.}

%such as prostate~\cite{wang2019deep}, the heart \cite{degel2018domain}, and carotid plaque segmentation \cite{zhou2019u}. 
% Modality of imaging and its varieties
%Magnetic Resonance (MR) has become a tool of choice for the investigation of neuromuscular diseases and sport~\cite{pillen2008muscle,loram2006use,ogier2017individual} due to its ability to distinguish fat and muscle tissue. 
%However, due to the early onset of the DMD disease, patients are often children, for whom MR imaging is unpractical. 3D Ultrasound (US) and Ultrafast imaging are rapidly evolving \cite{liu2019deep} offering an inexpensive and portable alternative. 
%Currently, US is the only real-time volumetric imaging modality \cite{dunnhofer2020siam} clinically available, yet needs further clinical validation. In this study, we rely on imaging data acquired from 3D freehand ultrasound modality. 
%Beside our two application of interests, there is other interest in 3D ultrasound segmentation in many other clinical applications such as segmentation of prostate cancer \cite{wang2019deep}, left atrium segmentation \cite{degel2018domain}, and carotid plaque segmentation \cite{zhou2019u}. 

%The manual and automatic segmentation of muscles are recognized as challenging tasks\cite{zlateski2018importance} given the anatomical variability.
%high variability of shapes and relative positions between muscles. 

In spite of being very time consuming and operator dependent, \textcolor{black}{neuromuscular studies often rely} on fully manual segmentation of 3D anatomical structures \cite{crawford2017manually,morrow2016mri}. It is therefore essential to develop automatic segmentation \textcolor{black}{or propagation} methods to aid such studies. Automatic muscle segmentation task in \textcolor{black}{3D} US should address several challenges including the anatomical variability such as the lack of contrast or texture differences between individual muscles, as well the US modality challenges \textcolor{black}{such as missing boundary or inhomogeneous intensity distributions} \cite{wang2019deep}. 
%\textcolor{black}{Contrary, muscle propagation task from a given sub-volume is challenged by changes in position, shape and appearance of the muscle due to the physics of the US beam or due to probe shifts~\cite{dunnhofer2020siam}.}
%DM
\textcolor{black}{Propagating a sub-volume mask to fill an entire volume presents additional difficulties such as the changes in position, shape and appearance of the muscle due to the physics of the US beam or the probe's motion~\cite{dunnhofer2020siam}.}

%and segmentation \cite{xie2018breast}, 
%diagnosis of focal liver lesions \cite{schmauch2019diagnosis} 
%and in analysis and visualisation of deep cervical muscle structure
Deep learning-based methods have made successful progress in the analysis of ultrasound images and videos for fetal localization~\cite{7090943}, breast and liver lesions classification \cite{han2017deep,schmauch2019diagnosis}, cervical muscle segmentation\cite{cunningham2016real}, \textcolor{black}{landmark tracking in liver sequences \cite{gomariz2019siamese}, and knee cartilage tracking \cite{dunnhofer2020siam}}.
%
%
%The success is interpreted by the availability of large annotated datasets, for which experimentation and validation are well understood, but limited by the necessary of supervision. For instance, annotated patient datasets enable the use of supervised learning methods but they must first be labelled by clinical experts at scale. 
The success of such fully-supervised methods relies on the availability of large datasets requiring the annotation by clinical experts at scale. 
Annotating ultrasound images at scale is a non-trivial, expensive, \textcolor{black}{irreproducible} and time-consuming task, especially when dealing with 3D or sequential data.

The interest has recently shifted towards learning from a limited quantity of annotated data such as few-shot learning\cite{zhao2019data} or self-supervision\cite{zheng2018fast}. \textcolor{black}{Kotia\etal\cite{kotia2020few} discussed few shot learning for medical imaging domain and addressed the problem of dependency on the availability of training data.} To learn representations from unlabelled input data, self-learning methods commonly rely on auxiliary tasks, such as image reconstruction \cite{hervella2018retinal,gonzalezduque:hal-02734902} or context restoration \cite{chen2019self}. Self-supervision may also exploit pseudo-labelling \cite{petit2018handling} wherein unannotated data are relabelled and reused for fine-tuning.

In this paper, we propose a novel deep learning segmentation \textcolor{black}{and propagation} method for 3D US data, \textcolor{black}{which requires \textit{few-shot} expert annotated slices per 3D volume, on average 48 annotations out of 1400 slices, and leverages unannotated sub-volumes using sequential pseudo-labelling.} To produce a fast and accurate muscle segmentation, suitable for reliable volume computation, we design a minimal interactive setting. Explicitly, we ask the expert to provide as initialization the first \textcolor{black}{sub-volume} (\textit{i.e.}~\textcolor{black}{the muscle mask for three contiguous slices}). Similar to Wug\etal\cite{wug2018fast}, we leverage \textcolor{black}{the coherence of the 2D masks over the volume depth}, to propagate the reference mask \textcolor{black}{sub-}volume.

% we leverage the \textcolor{black}{spatial and depth deformations} (spatiotemporal coherence) over a sequence of sub-volumes, to propagate the reference mask \textcolor{black}{sub-}volume. %

\begin{figure}[ht]
 	\centering
 	\includegraphics[width=\linewidth]{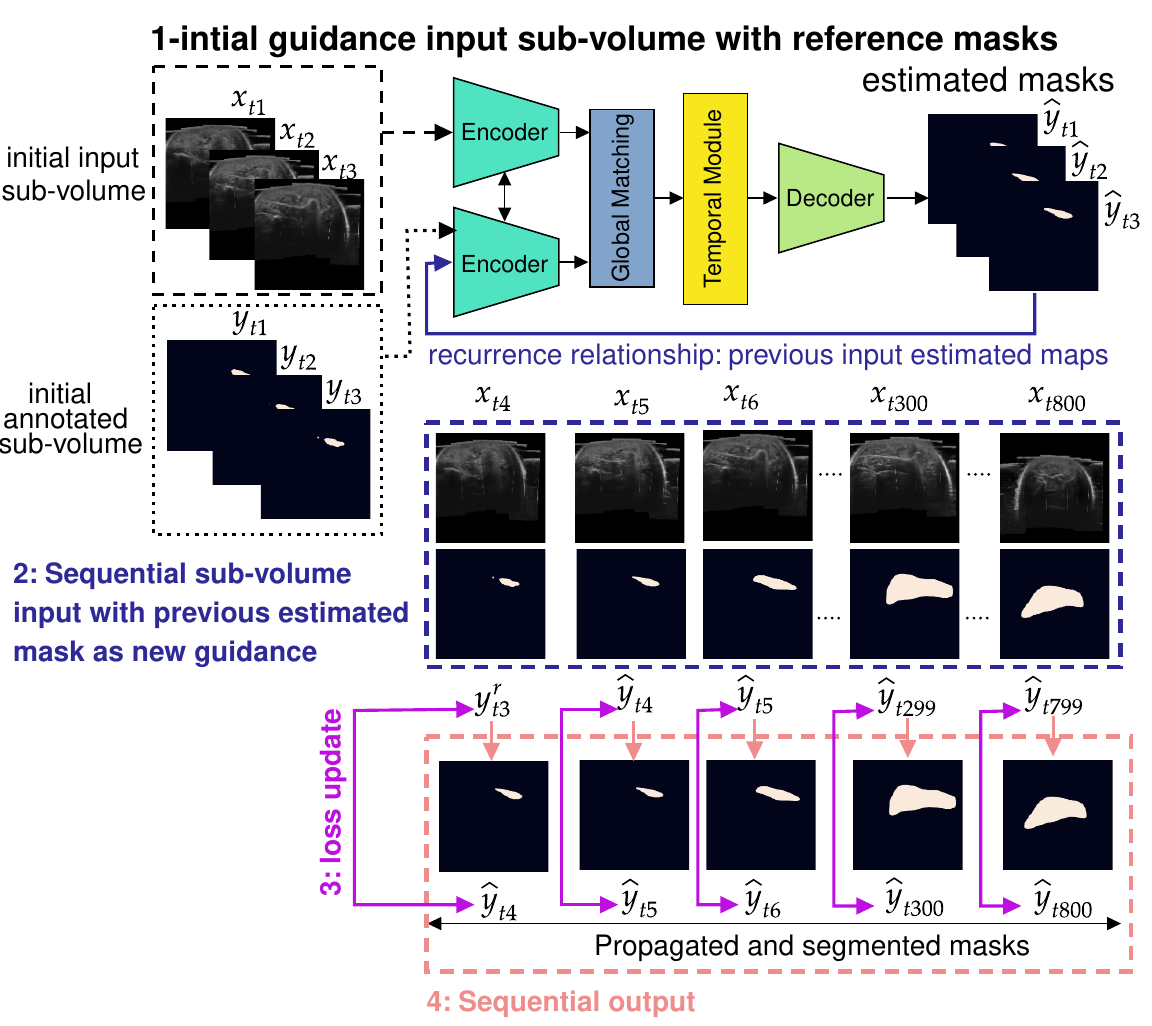}
    \caption{Mask propagation with the proposed end-to-end IFSS-Net. Shown here is an example propagation of a foreground mask from the first sub-volume annotated by an expert to the rest of the volume using a Siamese network.}
    \label{ModelOVerView}
\end{figure}

%In practice, we deploy a Siamese network to capture the representation of contiguous slices \textcolor{black}{within the current US sub-volume and the previous mask sub-volume prediction. The two encoder representations regarding the relative position of the muscle are matched with Global Feature Matching.} Capturing longer changes \textcolor{black}{and variations in positions and shape of muscle structure} are dealt with a Bidirectional Long Short Term Memory (Bi-CLSTM). 
%DM
In practice, we design a Siamese network to capture a common feature representation between ultrasound and mask sub-volumes. The reference can either come from an annotated part of the volume or from prior predictions. The common representation is further enforced with a global feature matching module~\cite{he2016identity}. The relationship between masks further away, encoding shape and structure changes at a larger scale are dealt with a Bidirectional Long Short Term Memory (Bi-CLSTM). An overview of our model \textcolor{black}{Interactive Few Shot Siamese Network} (IFSS-Net) is presented in Fig.~\ref{ModelOVerView}.

    % \caption{Mask propagation with end-to-end IFSS-Net. Shown here is an example propagation of a foreground mask from the first sub-volume expert annotation to the rest of the volume using Siamese network. A recurrence feedback loop for self-guidance is deployed. A sequential pseudo label strategy is proposed for loss update.}

%To guarantee the model convergence with limited annotated data, we propose a few-shot decremental strategy update of the objective function, in which we feed muscle volumes with annotated slices that are exponentially decreased and replaced by a relabelled annotations predicted from the last updated state of the model being trained. Finally, to handle the class-imbalance in the input training samples (i.e.,~fewer foreground pixels/voxels relative to the large number of background pixels/voxels in binary segmentation, as well as organs of varying sizes), we parameterize Tversky loss function to included two learned weights that penalize the false positive and the false negative. 

% To guarantee the model convergence with limited annotated data, we propose a decremental learning strategy. While we start feeding \textcolor{black}{sub-}volumes with labeled slices, we progressively \textcolor{black}{reduce the proportion of expert annotations \textcolor{black}{by} replacing missing annotations with predictions from the model.}

To guarantee the model convergence with limited annotated data, we propose a decremental learning strategy. While we start feeding \textcolor{black}{sub-}volumes with labeled slices, \textcolor{black}{we progressively reduce the proportion of expert annotations required for every new volume used for training, replacing them by predictions from the model.} Finally, to handle the class-imbalance between foreground and background pixels, we modify the Tversky loss\cite{salehi2017tversky} to adaptively learn the weights that penalize \textcolor{black}{\emph{False Positives} (FP) and \emph{False Negatives} (FN).}

We validate our approach for the segmentation, label propagation, and volume computation of the low-limb muscles, namely: the Gastrocnemius Medialis (GM), the Gastrocnemius Lateralis (GL), and the Soleus (SOL). We consider a dataset of $44$ subjects and $61600$ images, split into $29$ participants ($40600$ images) for training, 5 participants ($7000$ images) for validation and \textcolor{black}{a test set of 10 participants ($14000$ images)}. \textcolor{black}{We consider a fully-supervised setting to build an upper-bound reference. Then, we demonstrate our method's capability to learn from a few annotations under a simulated weakly-supervised regime, keeping only 3,5\% of the annotations (1420 images) and exploiting the remaining ``unannotated" images (39180 images) with sequential pseudo labeling.} The model's generalization was evaluated over a test set, resulting in a Dice score coefficient of over $95\%$.

% \textcolor{black}{For weak supervision setting}, we simulate having only $3.5\%$ ($1420$ images) \textcolor{black}{annotated} training set, while the rest of the unannotated images were exploited through sequential pseudo-labeling. 

\textcolor{black}{We first evaluate the model's performance when considering the volumetric segmentation task alone (without propagation). We refer to this model as Seg-Net-FS. This simplified approach allows for the comparison to other three state-of-the-art volumetric segmentation methods: 3D U-Net~\cite{cciccek20163d}, V-Net~\cite{milletari2016v} and DAF3D~\cite{wang2019deep}.} \textcolor{black}{We then consider the full IFFSS-Net, which combines the segmentation and propagation tasks, and evaluate its performance under both weak (3,5\% of the annotations) and full supervision (100\% of the annotations), comparing it to the recent mask-tracking method PG-Net~\cite{wug2018fast}. We also compare the performance of Seg-Net-FS with IFSS-Net to highlight the performance gain.} \textcolor{black}{Finally, to extensively assess the performance of our method, we designed several ablation setups to study the contribution of each proposed module. Finally, non-learning techniques were also considered for comparison}. \emph{The main contributions of this work are:}

% For \textcolor{black}{the} segmentation and label propagation task, we compare \textcolor{black}{the} IFSS-Net network under the weak ($3.5\%$ annotated data) and fully supervised ($100\%$ annotated data) settings with PG-Net~\cite{wug2018fast}. 

% We compare our network performances \textcolor{black}{for a volumetric segmentation task without propagation (we refer to this network as Seg-Net-FS) with state of the art methods such as} 3D U-Net~\cite{cciccek20163d}, \textcolor{black}{V-Net~\cite{milletari2016v} and DAF3D~\cite{wang2019deep}}. 

% train our ``IFSS-Net" network under the weak ($3.5~\%$ annotated data) and fully supervised ($100~\%$ annotated data) settings. \textcolor{black}{We compared the results to some biomedical volumetric segmentation based methods such as} ``3D U-Net" \textcolor{black}{\cite{cciccek20163d}, ``V-Net'' \cite{milletari2016v} and ``DAF3D" \cite{wang2019deep}, and also a recent propagation based-method ``PG-Net'' \cite{wug2018fast}.}

%In this paper, our main contributions are:

\begin{enumerate}
    \item A novel deep learning method for segmentation and muscle mask propagation in volumetric US data, towards accurate volume quantification.
    
    % \item A strategy to combine learning from few annotated 2D US slices with sequential pseudo-labeling of the unannotated slices.
    
    \item A sequential pseudo-labeling strategy for weak supervision to train from few annotated 2D US slices and exploit sequential unannotated slices.
    
    \item %Modelling the bidirectional spatiotemporal links to adapt to complex muscle shape and structure.
    A bidirectional spatiotemporal model to adapt to complex muscle shapes and textural changes over volume.
    % depth.
    % a longer range.
    
    % \item A decremental update of the objective function to guide the model convergence in the absence of large amounts of annotated data.
    
    \item A decremental update of the objective function to guide the model convergence in the absence of large amounts of annotated data and to \textcolor{black}{inducing a few-shot setting.}
    
    \item A parametric Tversky loss function that learns to adaptively penalize false positives and false negatives. 
    
    % \item Applying our model on 3D freehand US volumes and we asses the volume quantification measurements.
\end{enumerate}

\section{Related Work}
\label{sec:rw}

Our mask propagation problem is similar to that of \emph{Visual Object Tracking}. After specifying a target on the first frame, the goal is to track its evolution in shape and position over the following images. The task is achieved by modeling the spatiotemporal coherence between consecutive frames.

% In Computer Vision, Siamese networks are a common tracking tool as they are capable of learning similarities to identify related regions \textcolor{black}{between a reference frame and} contiguous frames\cite{tao2016siamese,bertinetto2016fully,wug2018fast,held2016learning,valmadre2017end}. When confronted with videos, these methods track either a bounding box around the object\cite{tao2016siamese,bertinetto2016fully}, or a reference segmentation mask~\cite{wug2018fast}. \textcolor{black}{In general Siamese network are composed of two twin encoders that share the same network parameters and configurations, permitting projection to the same feature space, where similarity matching can be applied.}
In Computer Vision, Siamese networks are a common tool for tracking. \textcolor{black}{Composed of twin encoders sharing the architecture and weight parameters, they project the images fed to the two branches to a common feature space where their similarity can be evaluated.} In this way, they are capable of learning similarities and identifying related regions among contiguous frames\cite{tao2016siamese,bertinetto2016fully,wug2018fast}. When confronted with videos, these methods track either a bounding box around the object\cite{tao2016siamese,bertinetto2016fully}, or a reference segmentation mask~\cite{wug2018fast}.

% Recently, 
Siamese networks have also been successfully applied to medical images. \textcolor{black}{Gomariz\etal\cite{gomariz2019siamese} formulated the problem of landmark tracking in liver-ultrasound sequences, relying on a Siamese network to find similar regions and a temporal consistency module as a location prior.} Dunnhofer\etal\cite{dunnhofer2020siam} used a Siamese for knee cartilage tracking in ultrasound images. Their model accepts as input a target image and a manual bounding box limiting the search area, and easing the segmentation task. \textcolor{black}{Li\etal\cite{li2020siamese} evaluate disease severity at single time points using medical image and modelled change over time between longitudinal patient visits on a continuous spectrum using Siamese network. Panteli\etal\cite{panteli2020siamese} used Siamese for tracking and modelling the movement behaviour of biological cells in video sequences.} We formulate our problem as a joint task of segmentation and propagation from reference masks and we avoid costly manual priors (\textit{i.e.}~bounding boxes or landmarks). Our work builds upon the Siamese architecture PG-Net by Wug\etal\cite{wug2018fast}, which considers both the detection and propagation of a target object in motion. While PG-Net produces sharp masks, it leads to unsmooth temporal transitions, which limit its performance when applied to sequential ultrasound data. PG-Net considers only the spatial information within a 2D image before propagating over time with a Recurrent Neural Network (RNN). To enforce mask smoothness, we design a recurrence relationship that connects predictions over time, similar to Hu\etal\cite{hu2017maskrnn} and Perazzi\etal~\cite{perazzi2017learning}. Such recurrence enables refining previous mask sub-volumes when making new predictions. Also, as in Khoreva\etal~\cite{khoreva2017lucid}, who consider future pixels, we model the muscle pixels in the past and future slices by integrating a Bidirectional Convolutional Long-Short-Term Memory (BiCLSTM) \cite{liu2017bidirectional}. With the recurrence relationships and the BiCLSTM module, we effectively enforce temporal smoothness while taking full advantage of the muscle changes along the volume.

To reinforce the learning of the local deformation patterns in image space and time for sequences within each US sub-volume (instead of only spatial as in \cite{wug2018fast}), we introduce Atrous Separable Convolutions (ASC)~\cite{chen2018encoder} into our model. 3D ASC differs from the typical 3D convolutional operator by an adaptable dilation rate that adjusts the filter's field-of-view. Thereby, 3D ASC captures contextual information at multiple scales. However, ASC may produce less sharp masks at the boundaries. Prior work has handled this issue with auxiliary refinement \cite{pinheiro2016learning} or reconstruction tasks \cite{gonzalezduque:hal-02734902}. Herein, we rely on a series of 3D ASC connected in a recurrent fashion to interpret the full context, while propagating contextual information in the bidirectional $z$-direction with the BiCLSTM.

\emph{Pseudo Labelling} is a semi-supervised strategy to cope with the difficulty of collecting annotations for large datasets. The strategy\cite{lee2013pseudo} \textcolor{black}{consists of two \textit{separated stages}: training over labeled data and using the predictions (pseudo-labels) of a deep network over the initial unlabeled data points to retrain the model.} Pseudo-labeling has been applied to classification and segmentation problems \cite{enguehard2019semi,wang2016cost,petit2018handling}, as well for correcting noisy labels in the context of active learning~\cite{lin2016re}. In this paper, we focus on training a segmentation model at the lowest annotation cost while leveraging pseudo-labeling on the high amount of unannotated slices to refine our propagation model. \textcolor{black}{Unlike prior work, we do not split learning into two separate stages. Instead, we propose a new \textit{continuous and sequential pseudo-labeling} scheme for volumetric data, taking advantage of the spatiotemporal smoothness between slices.} Our strategy starts from an annotated 3D US sub-volume sampled from the full volume and as we propagate to unlabeled sub-volume, we adapt the objective function to compare the current time-step prediction with its previous  time-step prediction, when no annotated data is available. \textcolor{black}{The advantage of sequential pseudo labeling is to keep a continuous gradient flow during training, under the assumption that contiguous slices are very similar/show very similar information.}

\textcolor{black}{The current state-of-the-art in volumetric segmentation relies on deep learning approaches and mainly on extensions of the U-Net architecture~\cite{ronneberger2015u} to 3D, such as the 3D U-Net~\cite{cciccek20163d} or the V-Net~\cite{milletari2016v}. Hesamian\etal\cite{Hesamian2019deep} presented a survey of popular methods that have employed deep-learning techniques for medical image segmentation and summarizes the most common challenges and some of the existent solutions.}

\textcolor{black}{Recent methods~\cite{roth2018towards,wang2019deep} have addressed the memory issues associated with 3D operations and focus on providing a full-segmentation mask for the whole volume at once. Roth\etal\cite{roth2018towards} replaced the concatenation layers of the 3D-UNet skip connections with summation layers and used a multi-GPU processor to segment full pancreas volumes. Similar to Roth\etal\cite{roth2018towards}, we rely on summation layers to reduce the total number of parameters. However, instead of processing whole volumes at once, we process sub-volumes sequentially, making it possible to segment large volumes with a single GPU. To overcome potential discontinuity artifacts induced by the sequential treatment, we employ a bidirectional spatiotemporal module (BiCLSTM).}

\textcolor{black}{Novikov\etal\cite{Novikov2018tmi} studied the segmentation of volumes by the sequential processing of 2D slices to generate a smooth prediction. The approach was based on a 2D UNet and two BiCLSTMs, one at the bottleneck and one at the last decoder layer. Although we also rely on a BiCLSTM, we seek to locally preserve the data's 3D nature; therefore, we do not rely on 2D slices but rather sub-volumes treated with 3D convolutions \cite{cciccek20163d}. The sequential processing of the sub-volumes is done through a BiCLSTM across the volume's depth, designed to capture texture and changes and deformations at a larger scale. Finally, in our BiCLSTM module design, we learn the weights for the forward CLSTM only and reuse them for the backward CLSTM to reduce the total number of parameters.}

\textcolor{black}{To handle the challenges of segmenting ultrasound images, Wang\etal\cite{wang2019deep} proposed a complex deep-learning architecture with several attention modules named DAF3D, which successfully segments prostate volumes. Although confronted with similar challenges, we take a different approach, opting for a light architecture and modeling the problem as the joint optimization of a segmentation and a propagation tasks. This multi-task approach is implemented through a Siamese Network~\cite{wug2018fast} instead of concurrent losses.}

\textcolor{black}{Common losses for image segmentation are the cross-entropy loss~\cite{ronneberger2015u}, the Dice score~\cite{milletari2016v}, or a combination of the two~\cite{wang2019deep}. However, these choices are not adapted to handle a large imbalance between the background and foreground classes~\cite{Hesamian2019deep}. To this end, Salehi \etal\cite{salehi2017tversky} proposed the Tversky loss, which generalizes the Dice and $F_{\beta}$ scores, achieving a trade-off between precision and recall by manually controlling the penalties for FPs and FNs with two weights $\alpha$ and $\beta$. We go one step further and propose to make the $\alpha$ and $\beta$ learnable parameters progressively optimized during training.}

\textcolor{black}{Dou\etal\cite{dou20173d} proposed a 3D deeply supervised network for volumetric scans. The method deploys a deep supervision to accelerate the optimization and boost the model performance, and proposes to refine the final contours with a Conditional Random Field (CRF). Despite the good reported performance, CRFs introduce an additional computational cost, and the deep supervision may be confronted with memory limitations for large volumes such as ours. Indeed, our volumes are significantly larger: $512\times512\times1400$ than those treated in all the above-cited works ($128\times128\times 3$ \cite{Novikov2018tmi},$160\times160\times72$ \cite{dou20173d}, $128\times132\times80$ \cite{wang2019deep}, ...), which has motivated our design choices in terms of a light architecture and sequential treatment of the volume.}

\textcolor{black}{In summary, in this work, we process a 3D US image, and predict a full volumetric mask without resizing the initial input. We overcome potential memory limitations by:}
\begin{enumerate}
\item \textcolor{black}{processing sequential sub-volumes and formulating the problem in terms of two joint segmentation and propagation tasks,}
    \item \textcolor{black}{building a lighter model that uses 3D atrous separable convolutions and shares weights between the forward and backward CLSTM module, and}
    \item \textcolor{black}{enforcing feature reuse and transfer using skip connections.}
\end{enumerate}

\textcolor{black}{We propose a parametric and learnable Tversy loss to overcome foreground/background pixels imbalance. We combine a modified Siamese architecture with the BiCLSTM module to model spatial and temporal coherence. Finally, we induce weak supervision through sequential pseudo labeling, and we improve the quality of the pseudo labels through a decremental update strategy, such that the training relies on very few annotated data.}

\section{Method}
\label{sec:m}

% \subsection{Overview}
\begin{figure}[ht]
 	\centering
 	\includegraphics[width=0.9\linewidth]{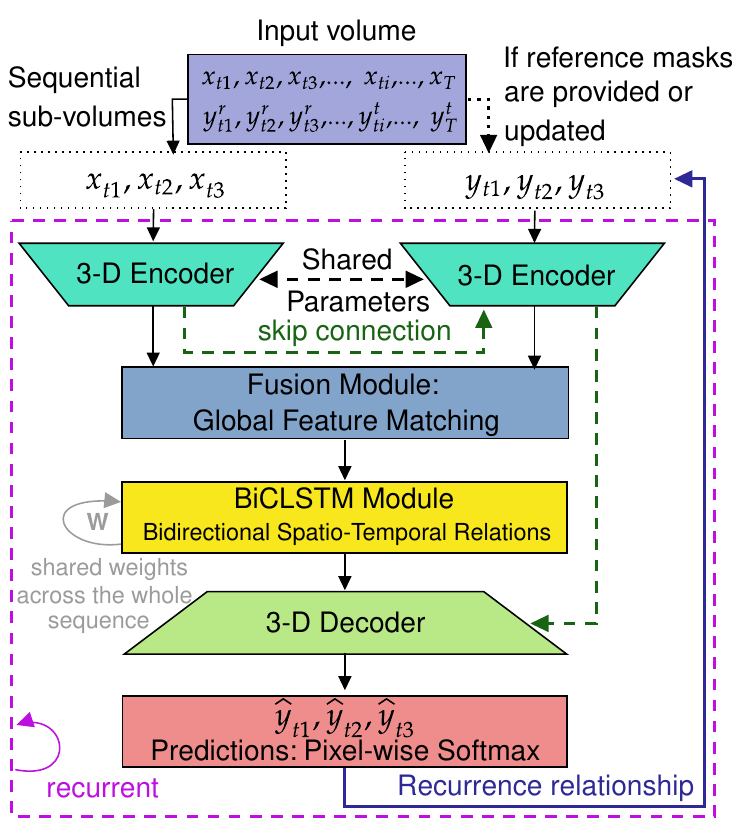}
    \caption{Our network architecture. The network consists of two twin encoders, a fusion module, a temporal module, and a decoder.}
    \label{GD}
\end{figure}

Consider a volumetric US volume as a stack of 2D slices along the $z-$direction. Given such a dataset, we \textcolor{black}{simulate} asking a clinical expert to provide the annotation of three 2D slices for the target muscle (SOL, GL or GM). The objective is to automatically segment the remainder of the volume by relying on additional poor/partial annotations. Since modeling the segmentation of the full 3D volume at once is intractable, we formulate the segmentation as the propagation of the provided annotations. We model the problem in a spatiotemporal fashion, where the temporal dimension is associated with the depth of the volume. We process the data with a sliding window handling one partial sub-volume at a time.

Observing that semantic features learned in an image segmentation task and appearance features learned in a similarity matching task complement each other, we combine a \textcolor{black}{baseline network 3D Seg-Net-FS we devised slightly based on 3D U-Net architecture \cite{cciccek20163d}} for feature extraction and segmentation with a Siamese tracking framework for muscle propagation. A general block diagram is demonstrated in Fig.~\ref{GD}. Our IFSS-Net architecture is composed of two twine 3D encoders with shared parameters. 
The first encoder captures feature representations from US sub-volumes, while the second captures representations from the segmentation masks. The fusion module for global feature matching fuses and matches the current muscle feature representation with the previous time-step masks representations. A memory module via Bi-CLSTM captures the spatial and depth changes from the previous, current and future slices. A 3D decoder maps the spatiotemporal information into a pixel-wise prediction. A recurrence feedback loop from the output to the mask encoding stream replaces the user interaction to keep it minimal. The whole model $\mathcal{M}$ is trained in a recurrent fashion using Truncated Back-propagation Through Time (TBPTT) \cite{werbos1990backpropagation}. Our objective function based on the parametric Tversky index is updated using labeled and pseudo-labeled data in a decremental fashion over upcoming subjects of the training dataset. The process ensures that the model has firstly learned proper and relevant target muscle features from the annotated set, before adding possibly noisy pseudo-labels. 
% Method overview
IFSS-Net firstly receives as an input a reference of US \emph{sub-volume} images with its annotated sub-volume masks. This step helps the IFSS-Net to discover the voxels locations of the target muscle to be localized, by matching the spatial features (muscle textures and patterns learned from spatial 2D slice) and depth deformation features (muscle shape \textcolor{black}{changes observed} over the depth direction, which we refer to as temporal information) at the reference US sub-volume. Then, IFSS-Net sequentially receives the subsequent sub-volume with their previous time-step mask pseudo labels. The recurrence relationship promotes the propagation of the previous sub-volume mask estimation to the current target muscle. A cooperating \emph{Bi-CLSTM module} allows \textcolor{black}{the capture of} both the appearance and the depth deformations, \textcolor{black}{thus promoting spatiotemporal consistency, while keeping relevant information through the gating mechanism} from each sequential sub-volume. 
Thereby, Bi-CLSTM is suitable for refining the mask propagation process using global contexts, \textit{i.e.} the entire volume. In this way, the proposed IFSS-Net automatically segments the target in every subsequent US sub-volume given the initial reference sub-volume.

% Removed to gain some space:
% In the following, we first formalize the problem, then we introduce the network architecture and some implementation details. Finally, we discuss our training procedure for updating the network parameters with few shot annotated samples. 

\subsection{Problem Formulation}

The dataset used for this work is composed of 3D ultrasound images of low limb muscles $\vect{\mathcal{V}}$ and their respective annotated masks $\vect{\mathcal{Y}}$ for SOL, GL and GM. We denote this data as $D=\{\vect{\mathcal{V}}_{i}, \vect{\mathcal{Y}}_{i}\}_{i=1}^{n}$, where $n$ is the number of patients. Each pair $(\vect{\mathcal{V}}_{i}, \vect{\mathcal{Y}}_{i})$ represents an ordered sequence of $T$ stacked 2D gray-scale US slices and their stacked annotated binary masks $\vect{\mathcal{Y}}_{i} = \{\vect{y}_{i}^{sol},\vect{y}_{i}^{gl},\vect{y}_{i}^{gm}\}$ indicating the localization of the muscles. The depth of the volume is denoted as $T\in \mathbb{N}$, being variable among different patients and muscles. Hence,$\vect{\mathcal{V}}_{i}$ can be expressed as $\{x_{1},...,x_{t},...x_{T}\} \in \mathbb{R}^{T \times 512 \times 512 \times 1}$ and $\vect{y}_{i}$ for a certain muscle (\textit{e.g.}~$\vect{y}_{i}^{sol}$) can be expressed as $\{y_{1},...,y_{t},...y_{T}\} \in \{0,1\}^{T \times 512 \times 512 \times 2}$, \textcolor{black}{with $2$ representing the foreground and the background channels.}

\begin{figure}[ht]
 	\centering
 	\includegraphics[width=0.9\linewidth]{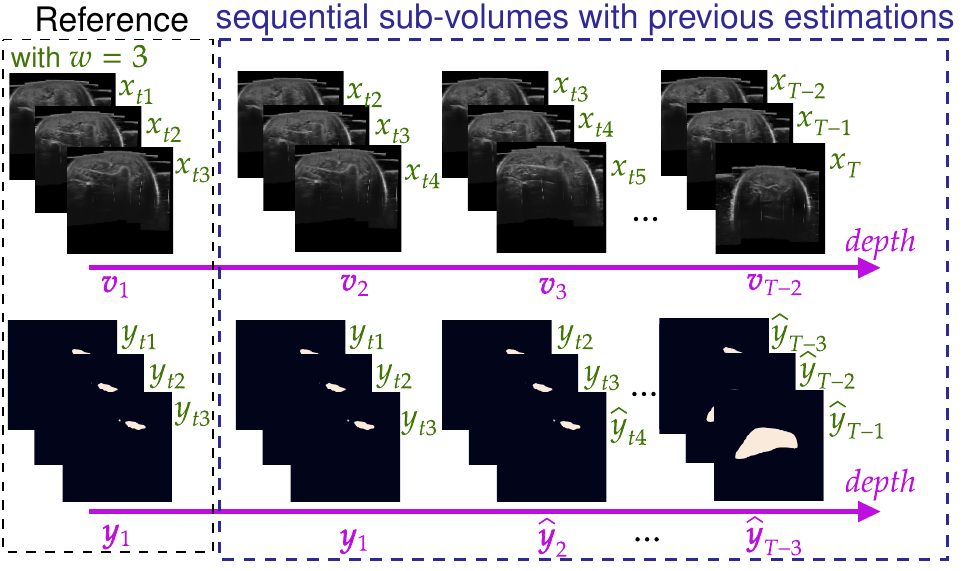}
    \caption{Sequential input of sub-volume with their corresponding previous time-step estimated masks. We can think about it as 3D+time.}
    % \textcolor{black}{of SOL muscle}
    % , where time represent the original volume depth
    \label{NetworkInput}
\end{figure}

\emph{The input:} we sample from a full volume $\vect{\mathcal{V}_{i}}$ a set of sub-volumes $\vect{v}_{i}$ by rolling a sliding window of size $w$ with \textcolor{black}{a} step size of $1$. Thereby, out of $T$ 2D slices, we create a new set of $T-w+1$ overlapped sub-volumes $\vect{\mathcal{V}}_{i}^{in}$, where $\vect{\mathcal{V}}_{i}^{in} = \{\vect{v}_{i,1},\vect{v}_{i,2},...,\vect{v}_{i,t=k},...,\vect{v}_{i,T-w+1}\}$. At a given time-step $t=k$, with time-step representing the index of a sub-volume corresponding to its depth in the $z$-stack, a sub-volume $\vect{v}_{i,k}$ is then composed of $w$ 2D US slices $\{x_{i,k},x_{i,k+1},x_{i,k+2},..,x_{i,k+w-1}\}$. The corresponding previous estimated sub-volume masks at time-step $t=k-1$ is $\vect{\hat y}_{i,k-1}$, composed of $\{{\hat y}_{i,k-1}, {\hat y}_{i,k}, {\hat y}_{i,k+1}, ..., {\hat y}_{i,k+w-2}\}$. Therefore,  $\vect{\mathcal{V}}_{i}^{in} = \{\vect{v}_{i,1},...,\vect{v}_{i,4},...,\vect{v}_{i,t=k},...,\vect{v}_{i,T-w+1}\}$ and $\vect{\mathcal{\hat Y}}_{i}^{in} = \{\vect{y}_{i,1},...,\vect{\hat y}_{i,3},...,\vect{\hat y}_{i,t=k-1},...,\vect{\hat y}_{i,T-w}\}$ are fed sequentially to the network. Fig.~\ref{NetworkInput} demonstrates an input example of sub-volumes composed of $w$-stacked images with overlap.

% Similarly, $\vect{\mathcal{Y}}_{i}^{in}$ is sampled. For the first step $t=1$, a reference mask map $\vect{y}_{i,t=1}^{ref} = \{y_{i,1},y_{i,2},y_{i,3}\}$ is provided by an expert along their corresponding three stacked images $\vect{v}_{i,t=1}^{ref}= \{x_{i,1},x_{i,2},x_{i,3}\}$. From $t=2$ to $t=T-2$, we consider $\vect{v}_{i,t}$ as a target volumes $\vect{v}_{i,t}^{tar}$, which we want to produce their corresponding masks $\vect{\hat y}_{i,t}^{tar}$ by propagating the reference mask $\vect{y}_{i,t=1}^{ref}$.

%\emph{The model:} we can think of our model $\mathcal{M}$ as a local spatiotemporal feature learning from 3D data for segmenting muscles and global spatiotemporal feature learning for propagating estimated masks through time.
\emph{The model:} we can think of our model $\mathcal{M}$ as simultaneously learning of local spatiotemporal features from 3D data for segmenting muscles, and global spatiotemporal features for propagating the estimated masks through depth direction.

% $\mathcal{M}$~takes~$\vect{\mathcal{V}}^{in}$~and~$\vect{\hat \mathcal{Y}}^{in}$~in a recurrence manner. 

\emph{The output:} for each sub-volume $\vect{v}_{i}$ in a set $\vect{\mathcal{V}}^{in}$, the estimated sub-volume masks $ \vect{\mathcal{\hat Y}^{in}} = \{ \vect{y}_{i,t=1},\vect{\hat y}_{i,t=2},...,\vect{\hat y}_{i,t=k},...,\vect{\hat y}_{i,t=T-w}\}$ are generated and fed back to the input. During the training stage, the set of sub-volume estimations $\vect{\mathcal{\hat Y}}^{in}$ are considered to update the loss function. During inference, only the first three annotated masks are provided by \textcolor{black}{simulating the expert}. For clarity, we will drop the patient index $i$ in the rest of the paper.

% I comment this part as it is minor, just some details about inference stage. ok
% Then, from each sub-volume prediction at time-step $t=k$, $\vect{\hat y}_{i,t=k} = \{\hat y_{i,k},\hat y_{i,k+2},...,\hat y_{i,k+w}\}$ only the first prediction $\hat y_{i,k}$ is considered, as the set $\vect{\mathcal{\hat Y}^{in}}$ consider overlap information. For the last time-step $T-w$, all the $w$ predictions in $\vect{\hat y}_{i,T-w}$ are considered. Hence, we form $\vect{\mathcal{\hat Y}}_{i}$ to match the initial input volume $\vect{\mathcal{V}}_{i}$, which is a sequence of 2D masks of depth $T$ and not 3D masks. For clarity, we will drop the patient index $i$ in the rest of the paper.

% We stick to the functionality, any further details to be kept for implementation details
\subsection{Network Structure}

% The neural network architecture we propose takes the advantages of the 3D U-Net structure for learning visual and motion features typically used in biomedical semantic segmentation, two twins 3D encoders with shared parameters and global feature matching operation used in Siamese framework for tracking and propagation and the Bi-CLSTM module for considering the full bidirectional spatiotemporal relationships. The whole model $\mathcal{M}$ is trained in a recurrent fashion using Truncated Backpropagation Through Time (TBPTT) \cite{werbos1990backpropagation} as shown in Fig. \ref{GD}. Later on, we provide unfolded version of it to provide understanding about the objective loss update.

\subsubsection{Siamese 3D Encoders}
The first 3D Encoder (${E}_\phi(\theta)$) processes the sub-volumes $\vect{\mathcal{V}}^{in}$ sequentially by taking at each time-step $t$ a sub-volume $\vect{v}_{t} \in \mathbb{R}^{w \times 512 \times 512 \times 1}$ and modeling its local appearance and depth deformation simultaneously. Typically, 3D convolutional operators are more appropriate for learning to extract spatiotemporal features compared to 2D convolutional operators.
% as the latter lack of depth deformation modeling and they lose depth information of the input signal right after every convolution operation. Even though, 2D convolutional operators could be utilized recurrently, after the first convolution layer, accumulating depth information usually collapse completely \cite{tran2015learning}. 
 In this study, we mainly use 3D Atrous Separable Convolutions (3D ASC). 

Each of the encoders ${E}_\phi(\theta)$ and ${E}_\varphi(\theta)$ has the same configuration and shared weights $\theta$. During the training phase, weight updates are mirrored across both sub-networks. We extract the local spatiotemporal features encoded from the US sub-volume $\vect{v}_{t}$ at \textcolor{black}{a} certain time-step (depth) $t$ by ${E}_\phi(\theta)$. Then, those spatiotemporal features are aggregated to ${E}_\varphi(\theta)$ using skip connections~\cite{he2016identity}. The latter helps ${E}_\varphi(\theta)$ to update the encoders weights $\theta$ to a better state as it has already accessed the previous location of the estimated muscle $\vect{y}_{t-1}$. 
% provided by the recurrence relationship using the previous estimation 

Aggregating the information from  ${E}_\phi(\theta)$ to ${E}_\varphi(\theta)$ is original in the sense that it reduces the computational resources. For instance, Wug\etal~\cite{wug2018fast} needed to feed for each of the ${E}_\phi(\theta)$ and ${E}_\varphi(\theta)$ a reference and target images %both associated with the guidance and estimated mask 
concatenated at the channel axis. If we followed the same idea, %it means that 
we would have to feed IFSS-Net four sub-volumes. Instead, our model accepts only one sub-volume per stream while aggregating the information from one stream to the other.

The second advantage is: feeding to ${E}_\phi(\theta)$ only the US muscle sub-volume without giving yet any prior knowledge about the possible target muscle locations 
%using the previously estimated masks in the same stream, it 
pushes the weights $\theta$ over ${E}_\phi$ to learn to detect the local spatiotemporal information independently from any possible prior knowledge. Then, when ${E}_\varphi(\theta)$ receives the prior about the previous estimated sub-volume mask, it establishes a new representation regarding the possible current target muscle locations and it uses the aggregated spatiotemporal information from ${E}_\phi(\theta)$ allowing $\theta$ to update and refine the semantic similarity between the two streams representations. By sharing the weights, the two streams map their representation into the same feature space. 

% Our Siamese 3D Encoders help the model convergence quicker than typical supervised learning methods because the ${E}_\varphi(\theta)$ is affecting the ${E}_\phi(\theta)$ update to extract more reliable local spatiotemporal features as the area where the muscle might be located is either provided by experts or estimated and fed back to the network. 

\subsubsection{Fusion Module}
% A global feature matching adapted from Peng\etal~\cite{peng2017large} and Zheng\etal~\cite{zheng2018fast} is applied between the $\varphi$ and $\phi$ streams to localize the target muscle. 

\begin{figure}[!ht]
 	\centering
 	\includegraphics[width=\linewidth]{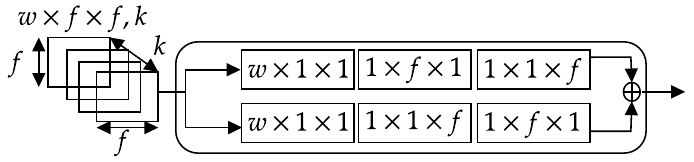}
    \caption{Global feature matching layer $g$ with enlarged receptive field by combining $[w~\times~1~\times~1]+[1~\times~f~\times~1]+[1~\times~1~\times~f]$ and $[w~\times~1~\times~1]+[1~\times~1~\times~f]+[1~\times~f~\times~1]$ convolution layers. The depth of the sub-volume is $w$ and $f$ corresponds to the bottleneck spatial size of \textcolor{black}{the} feature map with $k$ channels. The output of this layer, is processed further by one residual block~\cite{he2016identity}.}
    \label{GFM}
\end{figure}

% This operation allows to localize the target muscle in the current US sub-volume. 
A global feature matching layer $g$ adapted from Peng\etal\cite{peng2017large} and Wug\etal\cite{wug2018fast} is applied to the outputs of $E_\phi$ and $E_\varphi$ streams, $g = (E_\phi(\vect{v}_{t=k}),E_\varphi(\vect{\hat y}_{t=k-1}))$. The layer localizes and matches the appearance and depth deformation features correspondent to the current target muscle encoded by $E_\phi$ with the location features corresponding to the previous sub-volume predictions encoded by $E_\varphi$. \textcolor{black}{The} global feature matching operation is similar to applying a filter cross-correlation operation for capturing similarity between the two streams, as in\cite{dunnhofer2020siam}. However, our operation overcomes the locality of the convolution operation by efficiently enlarging the receptive field as shown in Fig.~\ref{GFM}.

% by by combining $[w~\times~1~\times~1]+[1~\times~f~\times~1]+[1~\times~1~\times~f]$ and $[w~\times~1~\times~1]+[1~\times~1~\times~f]+[1~\times~f~\times~1]$ convolution layers. Here, $w$ corresponds to the depth of the sub-volume and $f$ corresponds to the bottleneck spatial size of feature map with $k$ channels. The output of this convolution, is processed further by one residual block~\cite{he2016identity}. The final output of the fusion module $g$ has a shape of $1~\times~w~\times~f~\times~f~\times~k$.

\subsubsection{Bi-CLSTM Module}
To exploit the interslice and intraslice spatiotemporal muscle features information effectively, we introduce a temporal layer using Bi-CLSTM ($\psi$). Typical RNN's outputs are usually biased towards later time-steps, which reduces the effectiveness of propagating the relevant information over a full sequence of slices thus, resulting in unsmooth temporal predictions. This limitation is addressed \textcolor{black}{by} taking into account bidirectional spatial and depth changes.

A typical Bi-CLSTM layer consists of two sets of CLSTMs that extract features in two opposite directions, allowing the flow of information between the past $t-1$, the current $t$ and the future $t+1$ time-steps. One CLSTM operates from $g_{t-1}$ to $g_{t+1}$ while the other operates from $g_{t+1}$ to $g_{t-1}$. To reduce the number of learned parameters while still taking advantage of this module, we mimic the Siamese structure. Therefore, we consider one CLSTM layer in the forward direction and then we reuse the same CLSTM layer in the backward direction. Thereby, the same set of weights are forced to adapt to the appearance and depth deformations in both directions. As the parameters of the Bi-CLSTM module are shared over the entire volume of length $T$, %thereby, 
it first processes the local spatiotemporal feature representation obtained by $g$ and then it sequentially accesses and updates the shared weights using the rest of the \textcolor{black}{in}coming information at each time-step via special gates (input gate, forget gate, memory cell, output gate and hidden state). The outputs of the forward and backward CLSTM layers are merged using residual connections. 

% This layer allows us to handle the one time-step shift between the current US sub-volume processed by $E_\phi(\vect{v}_{t=k})$ and the previous softmax prediction $E_\varphi(\vect{\hat y}_{t=k-1})$. 
% Those gates process the bidirectional information to be kept, forget or passed. 

\subsubsection{3D Decoder}
The decoder $\mathcal{D}$ takes the output of the Bi-CLSTM module $\psi$ and also the features from the encoder $E_\varphi$ and then merges and aggregates them at different scales using a refinement module~\cite{pinheiro2016learning}. The refinement module considers the spatiotemporal information captured at the lower convolutional layers and the muscle-level knowledge in the upper convolutional layers, thus augmenting information in a top-down manner. $\mathcal{D}$ performs up-sampling operations and its final layer produces a high confidence prediction $\hat y_{t}$.

\subsection{Learning Stage}

Let $\vect{\hat y}$ and $\vect{y}$ be the set of predicted and ground truth binary labels respectively where $\vect{\hat y}$ and $\vect{y}$ $\in \mathbb{R}^{w\times512\times512\times2}$. The Dice similarity coefficient $D$ between two binary volume for segmentation evaluation is defined as:

\begin{equation}
    D(\vect{\hat y},\vect{y}) = \frac{2 |\vect{\hat y} \vect{y}|}{|\vect{\hat y}| + |\vect{y}|} = \frac{2 \sum_{i}^{N}\vect{\hat y}_{i}\vect{y}_{i}}{\sum_{i}^{N}\vect{\hat y}_{i}^{2} + \sum_{i}^{N}\vect{y}_{i}^{2}}
    \label{DiceSimCoefeq}
\end{equation}

\noindent where the sums run over the N pixels/voxels of the predicted binary segmentation sub-volume $\vect{\hat y}$ and the ground truth binary sub-volume $\vect{y}$. The objective loss \eqref{DiceSimCoefeq} if used in training, weighs FPs and FNs equally, which causes the learning process to get trapped in local minima of the loss function, yielding  predictions that are strongly biased towards the background. As a result, the foreground region is often partially detected.

To weigh FNs more than FPs since detecting small muscle is crucial, we propose \textcolor{black}{using} a loss layer based on the Tversky Index ($TI$) as in Salehi\etal~\cite{salehi2017tversky}. We extend Tversky loss ($1-TI$) to include learnable parameters $\alpha$ and $\beta$ that control the magnitude of penalties for FPs and FNs instead of tuning them manually. The Tversky Index is shown in \eqref{Tverskyindexeq}.

\begin{equation}
\resizebox{0.9\hsize}{!}{$
    TI(\vect{\hat y},\vect{y},\alpha,\beta) = \frac{\sum_{i}^{N}\vect{\hat y}_{0i}\vect{y}_{0i}}{\sum_{i}^{N}\vect{\hat y}_{0i}\vect{y}_{0i} + \alpha\sum_{i}^{N}\vect{\hat y}_{0i}\vect{y}_{1i} + \beta\sum_{i}^{N}\vect{\hat y}_{1i}\vect{y}_{0i}}$}
    \label{Tverskyindexeq}
\end{equation}

\noindent where $\vect{\hat y}_{0i}$ is the probability of voxel $i$ \textcolor{black}{being} a foreground of a target muscle and $\vect{\hat y}_{1i}$ is the probability of voxel $i$ \textcolor{black}{being} a background. The same applies to $\vect{y}_{0i}$ and $\vect{y}_{1i}$ respectively. Typically, we start with $\alpha$ and $\beta$ equal to $0.5$, which reduces \eqref{Tverskyindexeq} to \textcolor{black}{the} Dice coefficient as in \eqref{DiceSimCoefeq}. Then, $\alpha$ and $\beta$ gradually change their values, such that they always sum up to 1. In order to guarantee that $\alpha+\beta=1$, we apply a softmax function over those two parameters to generate a probability distribution. Moreover, we \textcolor{black}{take} advantage of the generalized Dice loss from Sudre\etal~\cite{sudre2017generalised} to accumulate the gradient computation over each sub-volume and we reformulate \textcolor{black}{the} Tversky Index as shown in \eqref{Tverskyindexeq2}. Therefore, we update the network weights in the right direction and by the right amount and we avoid the problem of vanishing gradient and unstable network behavior. 
%Thus, we reformulate Tversky Index as shown in \eqref{Tverskyindexeq2}.

% to consider the accumulation over in which instead of computing the Tversky loss of each time-step, the sum of the sum of the products over the sum of the weighted sums between the ground-truth and predicted probabilities is computed as shown in Equation \ref{Tverskyindexeq2}. 

% \begin{equation}
% \resizebox{.9\hsize}{!}{$
%     T(P,G,\alpha,\beta) = \frac{\sum_{j}^{3} \sum_{i}^{N}p_{0i}g_{0i}}{\sum_{j}^{3}\sum_{i}^{N}p_{0i}g_{0i} + \sum_{j}^{3}\alpha\sum_{i}^{N}p_{0i}g_{1i} + \sum_{j}^{3}\beta\sum_{i}^{N}p_{1i}g_{0i}}$}
%     \label{Tverskyindexeq2}
% \end{equation}

\begin{equation}
\resizebox{0.9\hsize}{!}{$
    TI(\vect{\hat y},\vect{y},\alpha,\beta) = \frac{\sum_{j}^{w}\sum_{i}^{N}\vect{\hat y}_{0i,j}\vect{y}_{0i,j}}{\sum_{j}^{w}\sum_{i}^{N}\vect{\hat y}_{0i,j}\vect{y}_{0i,j} + \sum_{j}^{w}\alpha\sum_{i}^{N}\vect{\hat y}_{0i,j}\vect{y}_{1i,j} + \sum_{j}^{w}\beta\sum_{i}^{N}\vect{\hat y}_{1i,j}\vect{y}_{0i,j}}$}
    \label{Tverskyindexeq2}
\end{equation}

\subsubsection{Full Supervised Baseline}

% \begin{figure}[ht]
%  	\centering
%  	\includegraphics[width=\linewidth]{./Figures/SupervisedUpdate.pdf}
%     \caption{Unfolding the recurrence relation into a full network showing the loss update at each time-step $t$ under full supervision. We accumulate the computation of Tversky Index $\{TI_{t}\}_{t=1}^{t=T-2}$ and then, compute the objective loss presented in Equation \ref{objectiveLossFullSupervision}.}
%     \label{SM}
% \end{figure}

% In this paper, we first train our network with full supervision mode over 29 patients. By that, 
We assume that for each input sub-volume in $\mathcal{V}_{i}^{in}$ for a certain patient $i$, its full sub-volume ground-truth $\mathcal{Y}_{i}^{in}$ is available. Our loss function $\mathcal{L}$ becomes:

\begin{equation}
    \mathcal{L} = 1- (\frac{1}{T-w+1})\sum_{t=1}^{T-w+1}(\frac{1}{w} \times TI(\vect{\hat y}_{t},\vect{y}_{t},\alpha,\beta)) + \lambda||\Omega||^{2}
    \label{objectiveLossFullSupervision}
\end{equation}

\noindent We include an $l_{2}$ regularization over the network parameters $\Omega$ with a decay rate equal to 0.00001 to prevent overfitting. Our loss accumulates all the local losses computed from $t=1$ to $T-w+1$. For US volumes, $T$ can be large, and accumulating the loss over long ranges might lead to %the problem of 
exploding gradients. To cope with this issue, we update the accumulated loss into two consecutive stages, 1) after it passes the first $\frac{T-w+1}{2}$ steps and 2) the last $\frac{T-w+1}{2}$ steps. $\mathcal{L}$ is optimized using ADAM updates\cite{kingma2014adam} with a scheduled learning rate that starts \textcolor{black}{at} 0.0001 and decreases to 0.00001 at the last two training epochs, to stabilize the weights updates. 

% we already said before
% TBPTT is used to update the network parameters in a recurrent fashion.

\subsubsection{Few-Shot Supervision Mode}

    % We accumulate the computation of Tversky Index $\{TI_{t}\}_{t=1}^{t=T-2}$.
\begin{figure}[ht]
 	\centering
 	\includegraphics[width=\linewidth]{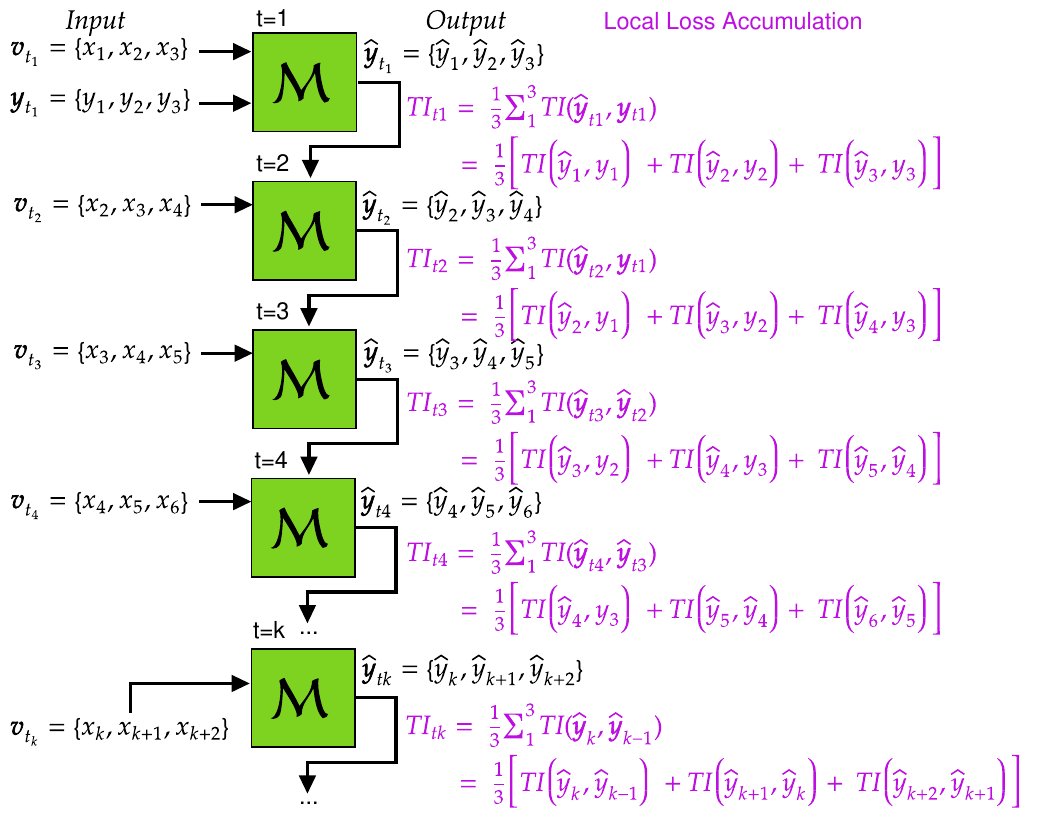}
    \caption{Unfolding the recurrence relation for the loss update at each time-step $t$ under few shot supervision with sequential pseudo labelling.
    }
    \label{supervisionWithFewShot}
\end{figure}

Consider for each sampled volume $\mathcal{V}_{i}^{in}$ for a patient $i$, a ground-truth $\mathcal{Y}_{i}^{in}$ that is sparsely annotated. The original volume $\mathcal{V}_{i}$ is composed of $T=1400$ 2D US slices. Let us assume the scenario were every 100 slices, a clinical expert provided only 3 consecutive annotations. Hence, out of $1400$ slices, only $3 \times 14 = 42$ annotations are provided. This amount represents around $3\%$ of the total volume and obviously it is not sufficient to train a neural network. In this paper, we provided few shot updates based on sequential pseudo-labeling. We relabel unannotated slices from the last updated state of $\mathcal{M}$ and consider them to update the loss function $\mathcal{L}$. We use manual annotations whenever they are provided to enhance the pseudo-labeling annotation. 

We demonstrate the \emph{few-shot update process }in Fig.~\ref{supervisionWithFewShot}. Let us assume that each sub-volume depth is $w=3$. To compute $TI$ at time-step $t=k$, we first produce the current estimated map $\vect{\hat y}_{t=k} = \{\hat y_{t=k}, \hat y_{t=k+1},\hat y_{t=k+2}\}$ at time-step $t=k$ and then we use the previous time-step pseudo-labelled estimation $\vect{\hat y}_{t=k-1}= \hat y_{t=k-1}, \hat y_{t=k}, \hat y_{t=k+1}$ at $t=k-1$ to update $TI$. We assume that such updates hold when the sequential spatiotemporal deformation over a sequence are smooth. Otherwise, the loss computation could become noisy and unpredictable.

\subsubsection{Decremental Learning Strategy}
A truly decremental deep learning approach for segmentation can be characterized by: (i) \textcolor{black}{the} ability to be trained from a flow of data, with \textcolor{black}{a} manually segmented mask disappearing in any order; (ii) achieving good segmentation performance; and (iii) \textcolor{black}{an} end-to-end learning mechanism to update the model and the feature representation jointly. In this paper, to benefit from the ``Few Shot Supervision Mode" for training $\mathcal{M}$. We implement a practical scenario that quickly converges $\mathcal{M}$ and produces less noisy pseudo-labeled annotations. Hence, instead of asking for 3 annotated masks every 100 slices, we ask for a gradual decrease of annotated masks ratio over patients. We still want to respect the $3.5\%$ annotation margin to train the whole model. Therefore, out of $n$ patients with a volume of $T$ stacked slices, we consider an exponential decay of the annotation \% over the patients respectively. For example, if $T=1400$, and $n=29$, \textcolor{black}{the number of image annotations required from the first participant to the last participant over a training set, will be as follow:$\{233,116,58,56,53,51,...,30,29,29,28,~\text{and}~28\}$}. The advantage of this decremental learning scheme is to provide the model with enough initial annotations to aid the process of detecting proper muscle features. Therefore, after the gradual decay of manual annotations, the model produces less noisy pseudo-labeled annotation. With such gradual decay, very few shot annotations can be utilized efficiently.

\subsubsection{Training and Implementation Details}

\emph{3D Siamese Encoder}. The twin encoders are composed of five stacked layers with a fixed filter size of $3\times3\times3$ and $[30,30,60,60,120]$ feature map. Each layer starts with a 3D ASC and followed by 3D max-pooling. The 3D ASC operation is applied at different rates $\{1,6,12,18\}$ which yield to a larger receptive field of $\{[3 \times 3 \times 3],[9 \times 9 \times 9],[15 \times 15 \times 15], [18 \times 18 \times 18]\}$ respectively. The obtained feature maps at different rates are concatenated along the channel axis and fed to the next layer. The final output of each of the two streams is $1\times 3 \times16\times16\times120$, where $1$ refer to the batch size, and in this study we process one patient at each iteration while the $3$ refer\textcolor{black}{s} to the sub-volume depth. A drop out layer is applied with a probability of 0.1. \emph{Bi-CLSTM Module.} Each of the CLSTM is composed of 120 feature maps with $\tanh$ activation function as it is bounded. Its convolutional filters are of size $3 \times 3$. The final output is of shape $1\times3\times16\times16\times120$. A dropout layer is applied with a probability of 0.4. \emph{Decoder}. It consists of five 3D up-convolutional layers. Each layer is composed of 3D transposed convolution and followed by refinement module for feature merging with the encoder $E_\varphi$ features and then dropout is applied with a probability of 0.1. The final layer produces a two-channel maps $\vect{\hat y}_{t}$ using $1 \times 1 \times 1$ convolution followed by pixel-wise softmax. \emph{Muscles Prediction.} Due to memory limitations, each muscle is trained independently. \textcolor{black}{Our architectures and modules are build upon: segmentation architecture~\cite{cciccek20163d}, propagation architecture~\cite{wug2018fast}, refinement module~\cite{pinheiro2016learning}, and fusion module~\cite{he2016identity}. The 3D ASC dilation rates are defined to form atrous spatial pyramid pooling~\cite{chen2018encoder}. The $\alpha$ and $\beta$ parameters of Tversky loss are learned adaptively. This leaves us only the two hyper-parameters for tuning over the validation set: the learning rate and the weight decay.}

% \textcolor{black}{Our initial hyper-parameters setting 

% were chosen on the bases of state-of-the-art studies discussed earlier for each module: segmentation architecture~\cite{cciccek20163d}, propagation architecture~\cite{wug2018fast}, 3D ASC parameters~\cite{chen2018encoder}, refinement module parameters~\cite{pinheiro2016learning}, and fusion module parameters~\cite{he2016identity}. Then they are adjusted to achieve the best performance the validation set.}

% and seen as a binary segmentation problem.

\section{Experimental Setup and Analysis}

% Out Previous Analysis  as it is
% \input{Sections/Comparison/PreviousComparison.tex}

% Setup of the Experimental Part.

\subsection{Low-limb Muscle Volume Dataset}

\begin{figure}[ht]
 	\centering
 	\includegraphics[width=0.9\linewidth]{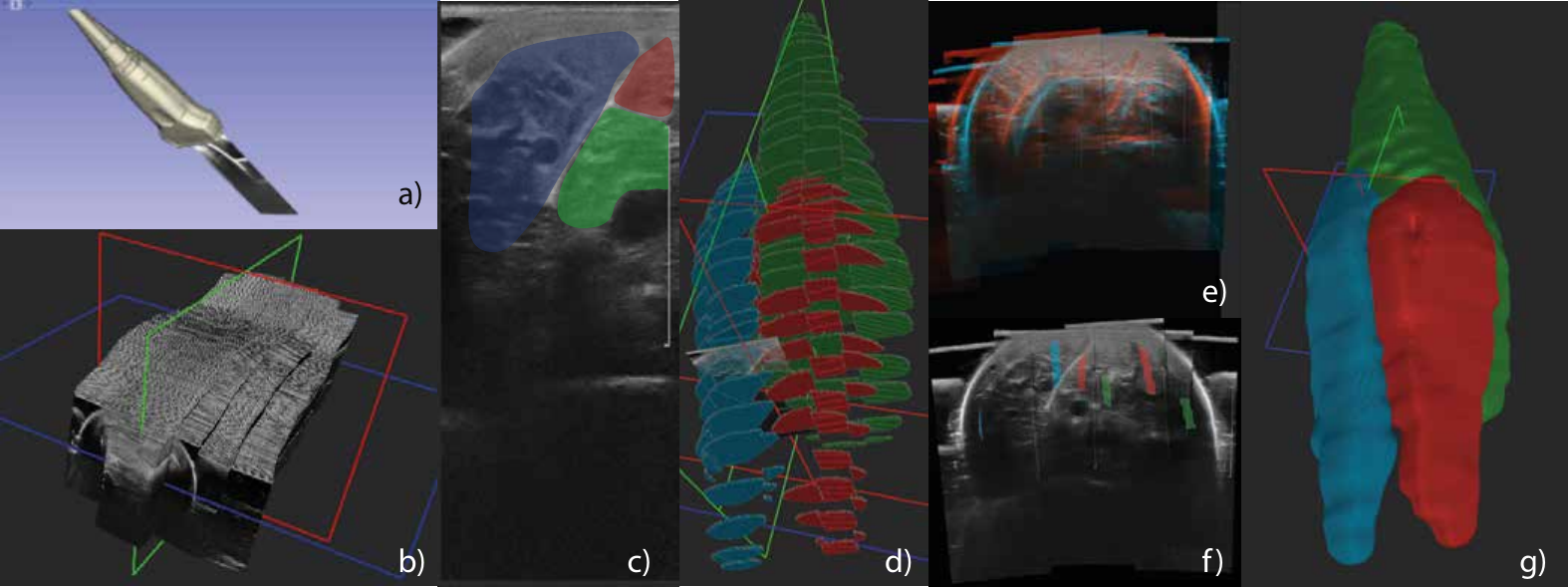}
    \caption{3D freehand US data: a)~Tracked probe b)~5 B-mode image sweeps c)~B-mode image with stradwin mask seeds, d)~GM in pink, GL in green and SOL in purple, e)~ Misaligned volumes, f)~Misaligned seeds, g)~Volume created after registration, label interpolation and polishing.}
    \label{datasetFig}
\end{figure}

%High resolution volumes
\subsubsection{Dataset acquisition}
In collaboration with Crouzier\etal\cite{crouzier2018neuromechanical}, 3D US recordings of 44 participants aged between 18 and 45 years old were collected. Participants were prone with their leg in a custom made bath to prevent pressure dependency in the measure. A total of 59 acquisitions were taken, 15 legs were recorded twice with different setting parameters to assure correct and complete visualization of the muscles. Four to six parallel sweeps were performed from the knee to the ankle~(Fig.~\ref{datasetFig}~b), under optical tracking of the probe~(Fig \ref{datasetFig} a). Images were recorded every 5 mm in low speed mode. High resolution 3D US volumes are compounded using the tracking matrices of the probe, filling a voxel grid of $564\times632\times1443\pm (49\times38\times207)$, with an average isotropic voxel spacing of $0.276993~mm^{3}~\pm 0.015~mm^{3}.$ 

%The sequence of 2-D B-mode US images was acquired with a Supersonix Ultrasound machine and a 40mm linear VERMON probe.

%Segmentation
\subsubsection{Mask annotations}
Annotation of GM, GL and SOL muscles were first approximated through interpolation of the seeds~(Fig \ref{datasetFig} d). The ``partial sparse seeds" were created over 2D B-mode US images using the Stradwin \cite{stradwin} software~(Fig \ref{datasetFig}c). After computing the error between the interpolated approximation and the fully manual slice by slice segmentation of 10 volumes (volumetric error of 4,17\%, Dice of 9\% and a mIoU of 14.3\%), we concluded interpolation alone was not reliable to train a learning method \cite{duque2020low}. Therefore, manual polishing by an expert was done over the interpolated volumes, leaving only 2 uncorrected and noisy approximations which we still use in the validation set. For patients with 2 recordings, GM and GL seeds are done over the first acquisition ($r_1$) while SOL seeds are done over the second one ($r_2$) with more gain and less frequency. Using 3D image-based rigid registration, we combine labels from different acquisitions~(Fig \ref{datasetFig} e-g). Then, we obtain a complete annotations of the 3 muscles over the volume reconstructed from $r_1$ acquisition.

%Split patient-wise
\subsubsection{Dataset splits}
In this study, our data split is done in a patient-wise manner. Out of the acquisitions of the 44 participants,  29 participants with  $29~\times~1400 = 40600~\text{images}$ are used for training. Those sequences are cropped and padded on volumes of size $512~\times~512~\times~1400$ to keep the voxel spacing unchanged. The GM, GL and SOL muscles are provided over a single volume. For the validation and the test set, we use 5 and 10 participants with 7000 and 14000 images.

\subsection{Evaluation Metrics}
To assess the segmentation outcome, we compute the Dice similarity coefficient (Dice) \textcolor{black}{as in \cite{milletari2016v}} and the mean Intersection over Union (mIoU) \textcolor{black}{as in \cite{cciccek20163d}}. Moreover, to quantify the smoothness and the surface error of the predicted binary volume, Hausdorff Distance (HDD in $mm$) and Average Surface Distance (ASD in $mm$) were computed  \textcolor{black}{as in \cite{wang2019deep}}. We report Precision (P) and Recall (R) values to show the trade-off when penalizing the FPs and the FNs and highlight the importance of using the Tversky index as in \textcolor{black}{\cite{salehi2017tversky}}. \textcolor{black}{The FPs are the pixels in the predicted mask while having zero intersection with the ground truth mask and vice versa for the FNs.} Finally, toward assessing volume measurements, we compute the volume (V) of each of the SOL, GL and GM muscles and report the percentage of error with respect to the ground-truth binary volume. To calculate the volume of the segmented muscles, the total number of pixels located inside the masks are added, considering the voxel spacing ($vs$). \textcolor{black}{The volumetric error percentage (VolErr) is then computed by dividing the absolute difference between the ground truth and predicted area volume over their average value.}

% \textcolor{black}{The FPs are the pixels in the predicted mask while having zero intersection with the ground truth mask and vice versa for the FNs
% while FNs are the pixels in the ground truth while having zero intersection in the predicted mask.} 

% \textcolor{black}{The FPs are the pixels in the predicted mask while having zero intersection with the ground truth mask and vice versa for the FNs
% while FNs are the pixels in the ground truth while having zero intersection in the predicted mask.} 

% To compute the \textit{Volume} from $\vect{Y}$ consisting of $T$ slices, \eqref{volcomp} is proposed. 

% \begin{equation}
%     Volume(\vect{Y}) =  \frac{(\sum_{t=1}^{T} \sum_{j=1}^{N-pixels} y_{j,t}) \times vs}{1000} cm^{3}
%     \label{volcomp}
% \end{equation}

\subsection{Methods Comparison}
\textcolor{black}{Our evaluation is divided into four parts, where we compare both qualitatively and quantitatively the performance of IFSS-Net. \emph{In the first part}, we evaluate the core architecture of our model for the automatic volumetric segmentation task only, trained in a fully supervised fashion. To this end, we simplify the full-model architecture in Fig.~\ref{GD} to ignore the propagation task by discarding: 1) the interactive setting, 2) the recurrence relationship, 3) the global feature matching and 4) the 3D twin Encoder $E_\varphi$. We refer to this baseline segmentation network as Seg-Net-FS, and we use to fairly compare our approach against other supervised volumetric segmentation methods such as 3D U-Net, V-Net and DAF3D networks, which are designed to perform segmentation on volumetric data in a fully supervised manner. \emph{In the second part,} we extended the Seg-Net-FS to IFSS-Net, by adding back all the discarded components, to jointly and reintegrating back the propagation task within the segmentation framework. Herein, we evaluate the performance of IFSS-Net: i) in fully supervised (IFSS-Net-FS) and weakly supervised (IFSS-Net-WS) manners, ii) against a state-of-the-art deep-learning propagation method (PG-Net) based on a Siamese Network~\cite{wug2018fast}, and iii) against the Seg-Net-FS to demonstrate the benefits of our IFSS-Net jointly considering the segmentation and propagation tasks. \emph{In the third part}, we perform a series of ablation studies to highlight the effectiveness of each of the modules defining our IFFS-Net.} \emph{In the fourth part}, we compare our methods to non-learning segmentation based methods.  We used the Slicer 3D open-source software~\cite{fedorov20123d} with built-in algorithms that propagate masks from initial reference annotations. We specifically use: Fill Between Slices (FBS), Grow from seeds (GFS) and Watershed (WS) methods. 

% We assessed the performances of segmentation and propagation networks by reporting: Dice Coefficient, mIoU, HDD, ASD, P and R. We also study the impact on the final volume computation (V).

% Our evaluation is divided into 2 parts. In the first part, we compare both qualitatively and quantitatively, the performance of IFSS-Net using the full supervision mode (IFSS-Net-FS) and the weak supervision mode (IFSS-Net-WS). Then, we compare our volume predictions against a state-of-the-art deep learning propagation method (PG-Net) based on a Siamese Network~\cite{wug2018fast}. Finally, we consider a 3D-Unet with full supervision as a baseline. All methods are reported over the validation and test data.
% %, for muscle segmentation in 3D Ultrasound scans. 
% In the second part of the evaluation, we compare our methods to non-learning segmentation based methods, both qualitatively and quantitatively. 
% We used the Slicer 3D open-source software~\cite{fedorov20123d} with built-in  algorithms that propagate masks from initial reference annotations. We specifically use: ``Fill Between Slices (FBS)", ``Grow from seeds (GFS)" and ``Watershed (WS)" methods.

% Include Full Supervision Mode without Interactive Setting and Comparison with 3D-Unet and DAF-3D
\subsubsection{Fully Supervised Volumetric Segmentation}

\textcolor{black}{We evaluate the blocks responsible for segmentation task alone. To this end we simplify the full model as described above, to build a baseline segmentation network (Seg-Net-FS). Then, we trained the Seg-Net-FS in a fully supervised manner and compared it to V-Net, 3D U-Net, and DAF3D, trained also under the same training protocol. We report the segmentation performances for the validation and test sets, in Tables \ref{tab:val} and  \ref{tab:test}. We also compare the number of parameters and run-time performance in Table \ref{tab:param}. Next, we discuss the advantages of Seg-Net-FS in terms of accuracy of prediction, memory cost and runtime.}

\begin{table}[h!]
  \begin{center}
    \caption{Average scores for the 3 low limb muscles over 5 validation participants.}
    \label{tab:val}
    \begin{tabular}{c|c|c|c|c} 
      \textbf{\backslashbox{Metrics}{Method}} & \textbf{3D U-Net} & \textbf{V-Net}& \textbf{DAF3D}& \textbf{Seg-Net-FS}\\
      \hline
      mIoU & \underline{0.578} & 0.491 & 0.468 & \textbf{0.785}\\
      Dice Coefficient & \underline{0.667} & 0.638 & 0.619&  \textbf{0.839}\\
      HDD & \underline{14.374} & 17.268 & 14.893& \textbf{6.595}\\
      ASD & 5.460 & 7.009 & \underline{3.576} & \textbf{3.391}\\
      Precision & 0.725 & 0.556 & \underline{0.651}& \textbf{0.853}\\
      Recall & 0.877 & 0.879 & \underline{0.656} &\textbf{0.907}\\
      VolErr & \underline{18.703} & 25.138 &  19.284 & \textbf{8.593}\\
    \end{tabular}
  \end{center}
\end{table}

\begin{table}[h!]
  \begin{center}
    \caption{Average scores for the 3 low limb muscles over 10 test participants.}
    \label{tab:test}
    \begin{tabular}{c|c|c|c|c} 
      \textbf{\backslashbox{Metrics}{Method}} & \textbf{3D U-Net} & \textbf{V-Net}& \textbf{DAF3D}& \textbf{Seg-Net-FS}\\
      \hline
      mIoU & \underline{0.778} & 0.562 &  0.544 & \textbf{0.833}\\
      Dice Coefficient & \underline{0.867}& 0.698 & 0.72 &  \textbf{0.894}\\
      HDD & \underline{6.646 }& 16.679 & 11.844 & \textbf{5.759}\\
      ASD & \underline{2.395} & 6.691& 3.028 & \textbf{2.014}\\
      Precision & 0.797 & 0.762& \underline{0.710} & \textbf{0.883}\\
      Recall & 0.973 & 0.897 &  \underline{0.746} & \textbf{0.919}\\
      VolErr & \underline{7.324} &  23.311  &18.558& \textbf{5.647}\\

    \end{tabular}
  \end{center}
\end{table}

\begin{table}[h!]
  \begin{center}
 \caption{Number of parameters and inference time for a volume of 1400 slices.}
    
    \label{tab:param}
    \resizebox{\columnwidth}{!}{%
    \begin{tabular}{c|c|c|c|c} 
      \textbf{\backslashbox{Metrics}{Method}} & \textbf{3D U-Net} & \textbf{V-Net}& \textbf{DAF3D}& \textbf{Seg-Net-FS}\\
      \hline
      Number of Parameters & 13,424,194 &\underline{ 8,410,184} & 28,918,390 & \textbf{4,129,026}\\
      Segmentation Time (s)  & 300.53 & \underline{230} & 290 &  \textbf{103.7}\\
    \end{tabular}}
  \end{center}
\end{table}

% Discussion on why we perform better rather than the discussion of the metrics, 

\begin{figure}[!ht]
 	\centering
 	\includegraphics[width=0.9\linewidth]{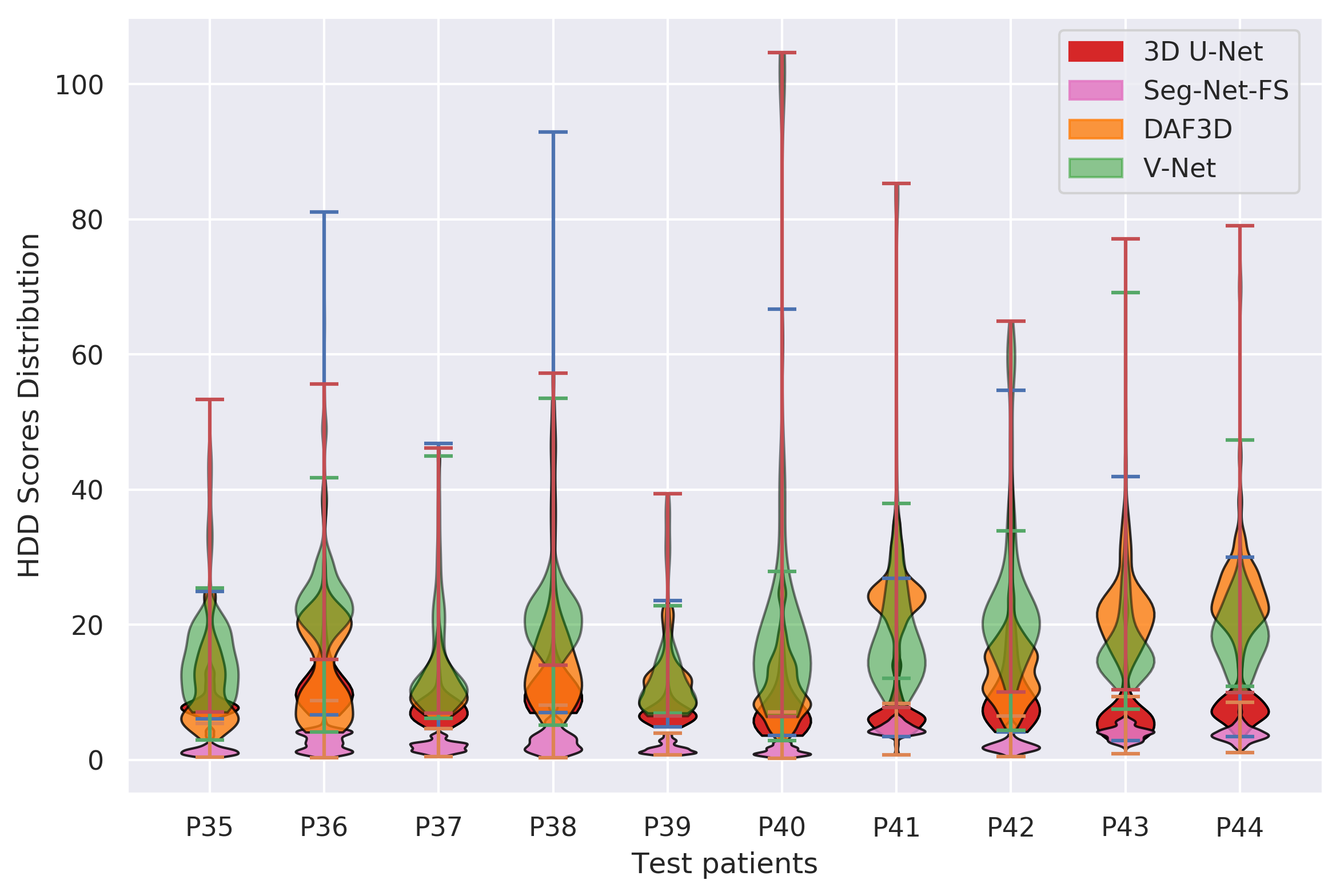}
    \caption{Hausdorff distance distribution over each of the test participants for different supervised segmentation methods.}
    \label{hddpart2}
\end{figure}

\textcolor{black}{The best performing method is Seg-Net-FS in term of all reported metrics. The second best performance in term of similarity metrics (Dice, mIoU and VolErr) is for 3D U-Net, while DAF3D is the second in term of prediction smoothness measures (HDD and ASD) as it achieved better precision-recall trade off. The poor performance of V-Net is due to the striding operation that leads to boundary information loss, as the 3D convolutional operation do not cover every pixel/voxel.} \textcolor{black}{In Fig.~\ref{hddpart2}, we show the distribution of HDD scores for each participant in the test set, highlighting which methods had difficulties in modelling the muscle shape deformations and variations over volume depth. We can see the HDD distribution for the Seg-Net-FS are tight and falls around a mean of $5.75~mm$. Similarly, the distribution of the scores for the 3D U-Net are tight but falls around a mean of $6.64~mm$. The HDD scores distribution for DAF3D varies across patients, in which we can observe that some patients their scores lies between first and third quartile such as  (``P36, P40, and P42"). This mean, that some slices where perfectly segmented (first quartile) while for the same patients, the rest of slices are badly segmented (third quartile). Finally, V-Net has a higher mean $16.65~mm$ and has more segmentation difficulties over the patients (``P36, P38, P42, and P44").} \textcolor{black}{Table \ref{tab:param} shows that our Seg-Net-FS provides a faster inference speed, the second best method is the V-net, followed by DAF3D and 3D U-Net, while utilizing less number of parameters due to our design choices presented in sections \ref{sec:rw} and \ref{sec:m}.}

\textcolor{black}{The reasons for the outstanding performance of our Seg-Net-FS are: 1) handling class imbalance using parametric Tversky loss, 2) capturing local deformation patterns from full resolution sub-volumes of shape $512\times512\times3$ without losing spatial information due to resizing of the volume to fit in memory as in DAF3D, 3D U-Net or V-Net, 3) handling longer muscle deformations over the entire volume using BiCLSTM module, 4) our 3D encoder and decoder are equipped with better 3D atrus separable convolution operators that extract multi-scale features with different dilation rates and 5) we better regularize our model with dropout and $l_{2}$ regularization.}

\textcolor{black}{\emph{Discussion:} In this work, we consider the time taken to segment an US volume and the average volumetric error between two annotators as a reference. On our dataset the two annotators on average spent around 2 hours to sparsely} \textcolor{black}{annotates 2D slices on each volume, followed by using some on the shelf interpolation methods to propagate their seed. Finally, they perform polishing for correction and to obtain the full volume annotation. The VolErr between them was reported as 3.5\%. Therefore, having 8.593\% and 5.647\% volumetric error over the validation and the test set in 103.7 second using Seg-Net-FS is a good progress. To further improve the Seg-Net-FS network performance, and to probably scale across new datasets, we propose to leverage sparse experts interactions. Our approach learns to propagate the expert's initial mask through the volume. Furthermore, it relies on a semi-supervised learning scheme intended to reduce the total number of manual annotations and polishing corrections.}

\subsubsection{Joint Segmentation and Propagation}
\textcolor{black}{From the analysis of Seg-Net-FS, we deduce how difficult it is for a strongly supervised network to cope with all the anatomical and ultrasound modality difficulties. Hence, to improve the segmentation performances, we jointly optimize for the segmentation and propagation tasks. At initialisation, the IFSS-Net uses a first annotated sub-volume to localise the target muscle. The localised muscle mask is then updated using the recurrence relationship to refine the segmentation of the current sub-volume. Finally, the BiCLSTM takes previous, current and future information  into account, to adapt to the global muscle mask deformations. As we demonstrate in the following experiments, these improvements help the IFSS-Net outperform the Seg-Net-FS.} \textcolor{black}{We compare the proposed model performance under weak (IFFS-Net-WS) and full supervision (IFFS-Net-FS), and compare it against the PG-Net propagation method. We also report the volumetric error to validate that our extension from Seg-Net-FS to IFSS-Net promotes a more accurate volume quantification.}

% We can make subsection dedicated to volume error fraction and some visual analysis
% Fig.~\ref{VolumeError}~shows the \textcolor{black}{average error (in percentage) of the computed muscle volumes starting from the predicted segmentation. }

\begin{figure}[!ht]
 	\centering
 	\includegraphics[width=\linewidth]{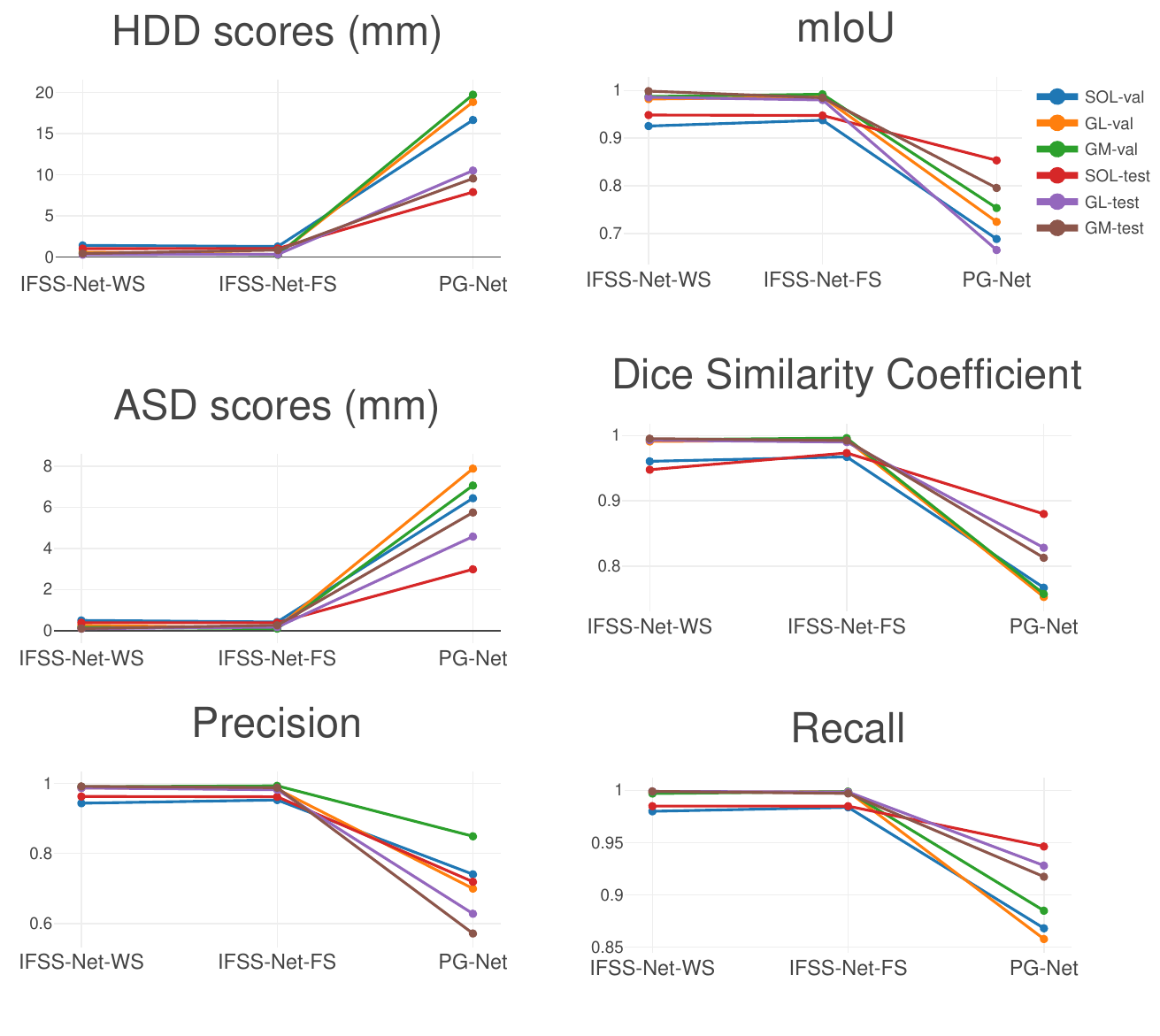}
    \caption{Overview of metrics for the different segmentation methods over validation and test sets.}
    \label{InteractiveSettingValidationTest}
\end{figure}

\begin{figure}[ht]
 	\centering
 	\includegraphics[width=0.85\linewidth]{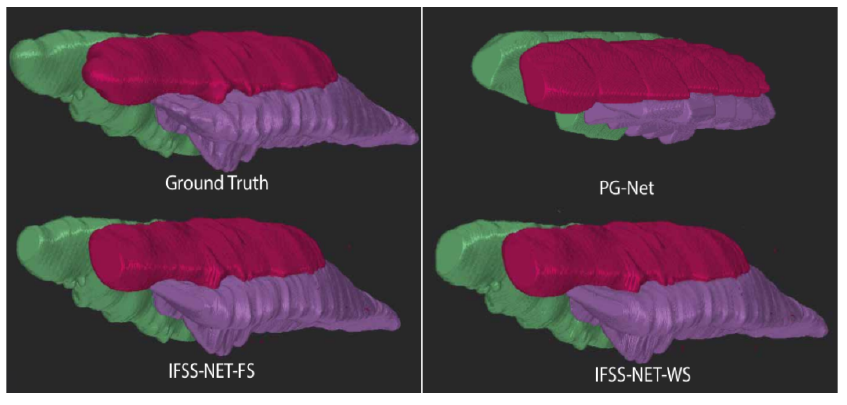}
    \caption{Predicted volume of GM (pink), GL (green) and SOL (violet) volumes over four methods for one test patient.}
    \label{qualitativeAnalysis}
\end{figure}

\begin{figure}[!ht]
 	\centering
 	\includegraphics[width=0.9\linewidth]{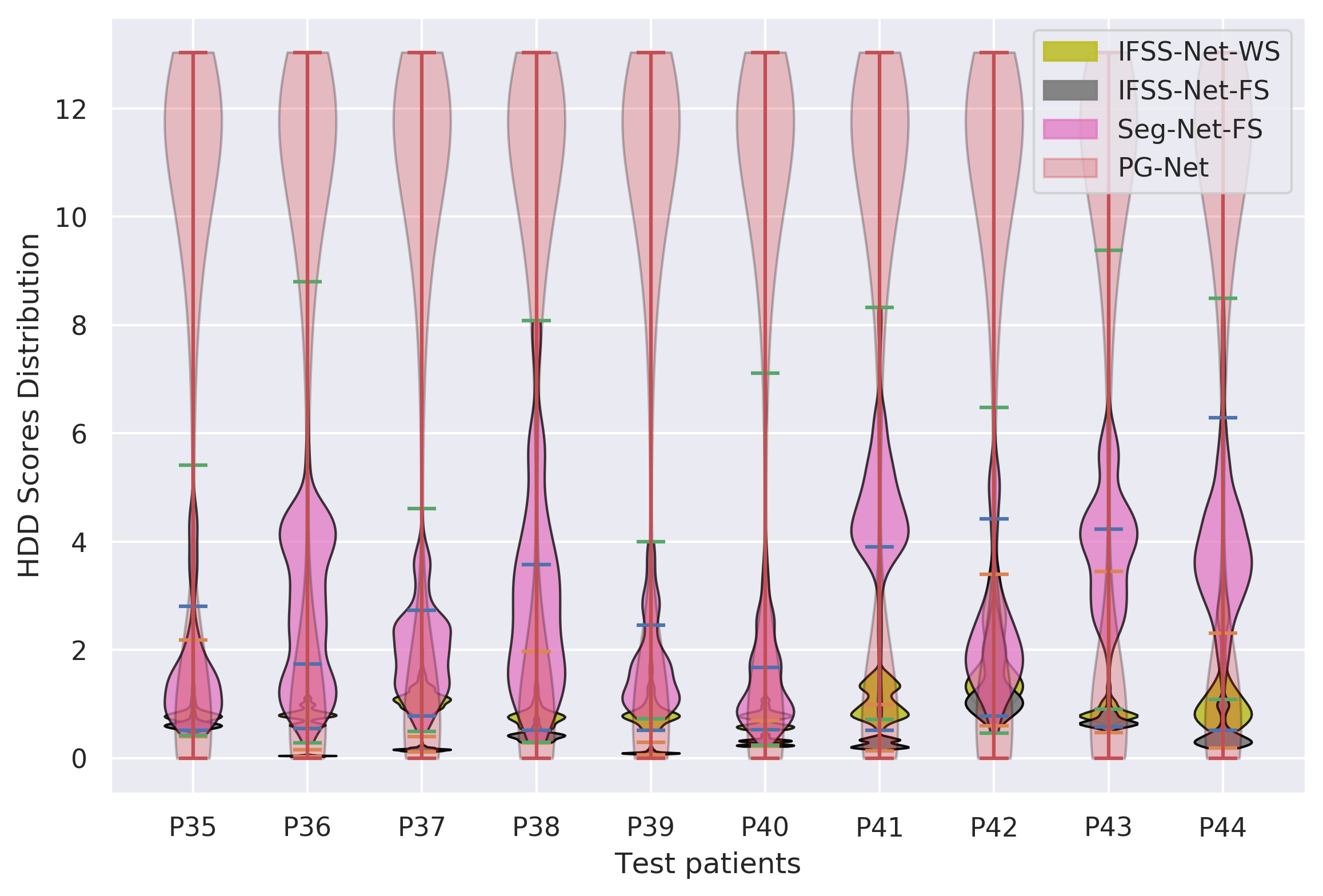}
    \caption{Hausdorff distance distribution over each of the test participant\textcolor{black}{s} for different propagation and segmentation methods.}
    \label{hddpart3}
\end{figure}

\textcolor{black}{In Fig.~\ref{InteractiveSettingValidationTest} we show\textcolor{black}{s} the obtained metrics for the three propagation methods over the validation and the test set.} \textcolor{black}{The best measures were achieved for the IFSS-Net-FS reporting over the three muscles an average of: $0.9872 \pm 0.0016$ Dice, $0.977 \pm 0.0047$ mIoU, $0.578 \pm 0.039~mm$ HDD and $0.197 \pm 0.026~mm$ ASD. 
Fig.~\ref{qualitativeAnalysis}~shows for the IFSS-Net-FS a very smooth prediction and a closer to the ground truth segmentation yielding to a small volumetric error with $1.129 \pm 0.045$ and $0.742 \pm 0.015$ as an average over the three muscles for the validation and the test set respectively. IFSS-Net-FS is followed by the IFSS-Net-WS, with $0.985 \pm 0.004$ Dice, $0.971 \pm 0.006$ mIoU, $0.728 \pm 0.084~mm$ HDD and $0.275 \pm 0.031~mm$ ASD. IFSS-Net-WS yields to an average \% of volume error of $1.495 \pm 0.07$ and $0.981 \pm 0.08$ over the validation and the test sets respectively. The weakly supervised prediction, as shown in Fig.~\ref{qualitativeAnalysis}, is also smooth and achieves competitive performance \textcolor{black}{for} IFSS-Net-FS. PG-Net falls to the third place in terms of Dice and mIoU measures.  As shown in Fig.~\ref{qualitativeAnalysis} it performs a segmentation similar to zero-order interpolation, yielding to an unsmooth volume surface prediction. \textcolor{black}{This is due to the PG-Net architecture using the RNN module, which requires flattening feature maps into vectors, thus loosing spatial structure.} Therefore, PG-Net reported high HDD $18.407 \pm 0.13~mm$ and $7.126 \pm 0.16~mm$ ASD scores, while yielding to $18.617 \pm 3.984$ \% of volume error. Regarding the P and R, we can see that our model maintains a proper trade-off, yielding a successful segmentation.} \textcolor{black}{In Fig.~\ref{hddpart3} we show the HDD distribution over the propagation networks and also we compares it with Seg-Net-FS. We can see that our extension from the volumetric segmentation (Seg-Net-FS) to segmentation with propagation (IFSS-Net) promotes a great reduction in the HDD scores, yielding to a smoother prediction while promoting accurate volume quantification. We can see that PG-Net has two set of quartiles, those concentrated below $4~mm$ which corresponds to the propagated masks that are close to the reference masks, and those above $8~mm$, which corresponds to the propagated masks that are far from the reference masks. Therefore, PG-Net have difficulties in propagating the reference mask in depth.}

%  reporting over the three muscles an average of:

\begin{figure}[ht]
 	\centering
 	\includegraphics[width=\linewidth]{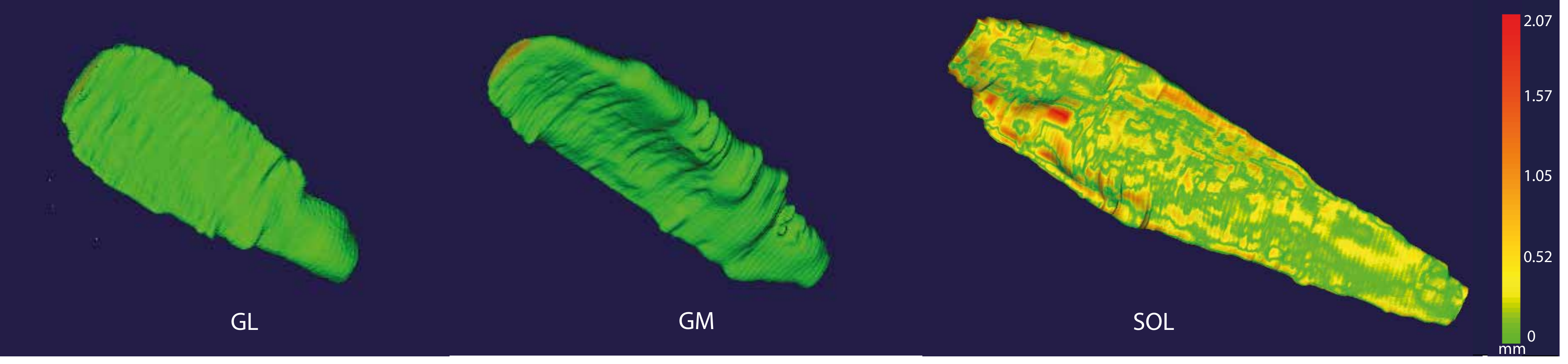}
    \caption{Surface error map for IFSS-Net-WS. The color bar indicate the surface error where the values go from $~0~mm$ until $2,07~mm$.}
    \label{surfaceDistance}
\end{figure}

\textcolor{black}{Regarding our IFSS-Net-WS, we visualize the surface error map over GL, GM and SOL to analyze if the muscles were equally difficult to segment. Fig.~\ref{surfaceDistance}~shows the distance to the ground truth in color for each muscle. We can see that SOL muscle is the hardest to segment. Highlighted sub-volume regions in red are for the most part explained by the poor quality of the US images in those regions. For GM and GL, the endpoints were harder to segment, while not surpassing $0,52~mm$ of distance error. This might be due to our model failing to propagate the masks until the muscle endpoints as the US sub-volume become noisy.}

% Commented to reduce space or to be commented
\begin{table}[h!]
  \begin{center}
    % \caption{Average volumetric error in \%, the total number of models parameters and the segmentation time in seconds taken to segment a volume of 1400 slices.}
    \caption{Average volumetric error in \%, Number of parameters and inference time for a volume of 1400 slices.}
    \label{tab:avrgVolumetricErrorParametersandTime}
    \resizebox{\columnwidth}{!}{%
    \begin{tabular}{c|c|c|c|c} 
      \textbf{Metrics} &\textbf{IFSS-Net-FS} & \textbf{IFSS-Net-WS}& \textbf{Seg-Net-FS}& \textbf{PG-Net}\\
      \hline
      Parameters & \underline{6,217,186} & 6,217,186  & \textbf{4,129,026} & 8,649,224\\
      Time &\underline{130} & 130 & \textbf{103.7} & 190.3\\
      VolErr & \textbf{1.23$\pm$0.46} & \underline{1.60$\pm$0.58} & 7.12$\pm$2.9 & 18.61$\pm$ 3.98\\

    \end{tabular}}
  \end{center}
\end{table}

In Table \ref{tab:avrgVolumetricErrorParametersandTime}, \textcolor{black}{we report the number of parameters of the new extended model and the time taken for the segmentation and} \textcolor{black}{propagation task. We also reported the average volumetric error \% over the validation and test set. We can see that our extension from Seg-Net-FS to IFSS-Net increased the inference time by $26.3$ seconds, due to the added components. Nevertheless, it promotes performance gain while significantly reduces the VolErr from $7.12\%$ to $1.23\%$ and $1.6\%$ and in full and weak supervision respectively.}

\textcolor{black}{\emph{Discussion:} While two expert annotators had achieved a VolErr of $3.5\%$, our IFSS-Net-WS achieved only $1.6\%$ with a faster performance of $130$ seconds. Obviously, optimizing two complementary segmentation and propagation tasks is important and beneficial for volumetric error reduction.}

% Ablation Studies.

\subsubsection{Ablation Analysis}
\textcolor{black}{To asses the performance gain brought by each of the modules of the proposed IFSS-Net, we designed six ablation experiments. The quality of the predictions is measured with average of the mIoU and the ASD scores over the validation and the test sets. We also report the time (in seconds) taken to propagate a reference sub-volume mask and obtain a binary mask prediction for the whole volume. For a qualitative analysis, we plot the ASD and the mIoU scores versus the slice index over a sub-volume $\mathcal{V}_{sub}$, using one of the validation patients. $\mathcal{V}_{sub}$ is composed of $720$ slices that cover the whole muscle from the starting point to the end. In the ablation study, we only consider the SOL muscle being the hardest muscle to segment.}

\begin{figure}[ht]
 	\centering
 	\includegraphics[width=\linewidth]{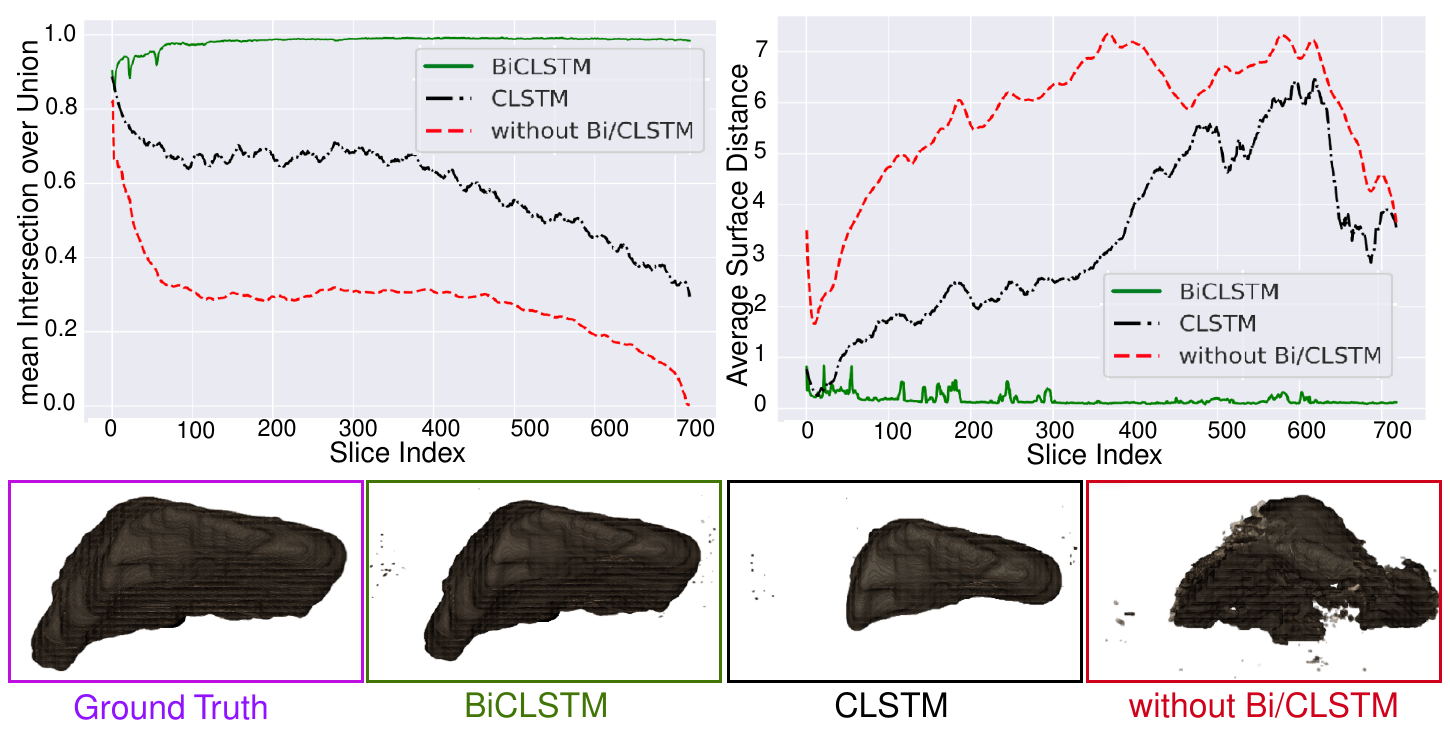}
    \caption{Ablation 1: network performance using different temporal modules. 
    \textbf{(Top)} mIoU and ASD scores over depth (slice index), to asses the propagation for each slice. \textbf{Bottom} Resulting rendered volumes for one subject.
    }
    \label{TM}
\end{figure}

\textcolor{black}{
\textit{Ablation 1: Temporal Module}. We change the BiCLSTM module by a one directional CLSTM module. Another ablation is conducted by discarding the temporal module and directly sending the fused global matching features to the 3D decoder for computing the final prediction. Table~\ref{tab:TM} reports the obtained scores and Fig.~\ref{TM} presents a qualitative visualisation over one validation sample. The BiCLSTM module considers the information from current, past, and future annotations helping the network learn how the SOL muscle appearance varies along depth and how to distinguish it from the other two (GL and GM) muscles. It maintains a fast response response (around 130 seconds) and yields a smooth prediction with a low ASD and high mIoU scores, as shown in Table~\ref{tab:TM}. When the BiCLSTM is replaced with the CLSTM module, the propagation is still smooth at the beginning but it gradually degrades (see Fig.~\ref{TM}). The mIoU decreases as the propagation starts to fade and the ASD score increases gradually since no future predictions or annotations are incorporated in the learning process. Finally, without the BiCLSTM or CLSTM modules, the network is not capable of producing consistent propagation as the SOL mask predictions leak to the surrounding GL and GM muscles.}
% The propagation time is around 130 seconds to propagate and segment the SOL muscle using our method, however, without the BiCLSTM module, it is reduced to 100 seconds.

\begin{table}[h!]
  \begin{center}
    \caption{Average ASD and mIoU scores and time to segment a volume.}
    \label{tab:TM}
    \begin{tabular}{c|c|c|c} 
      \textbf{\backslashbox{Metrics}{Method}} & \textbf{BiCLSTM} & \textbf{CLSTM}& \textbf{Without Bi/CLSTM}\\
      \hline
      ASD score & \textbf{0.171} & \underline{3.523} & 5.745 \\
      mIoU & \textbf{0.9827} & \underline{0.597 }& 0.282\\
      Propagation Time (s) & 130 & \underline{123} & \textbf{100}\\
    \end{tabular}
  \end{center}
\end{table}

\begin{figure}[ht]
 	\centering
 	\includegraphics[width=0.95\linewidth]{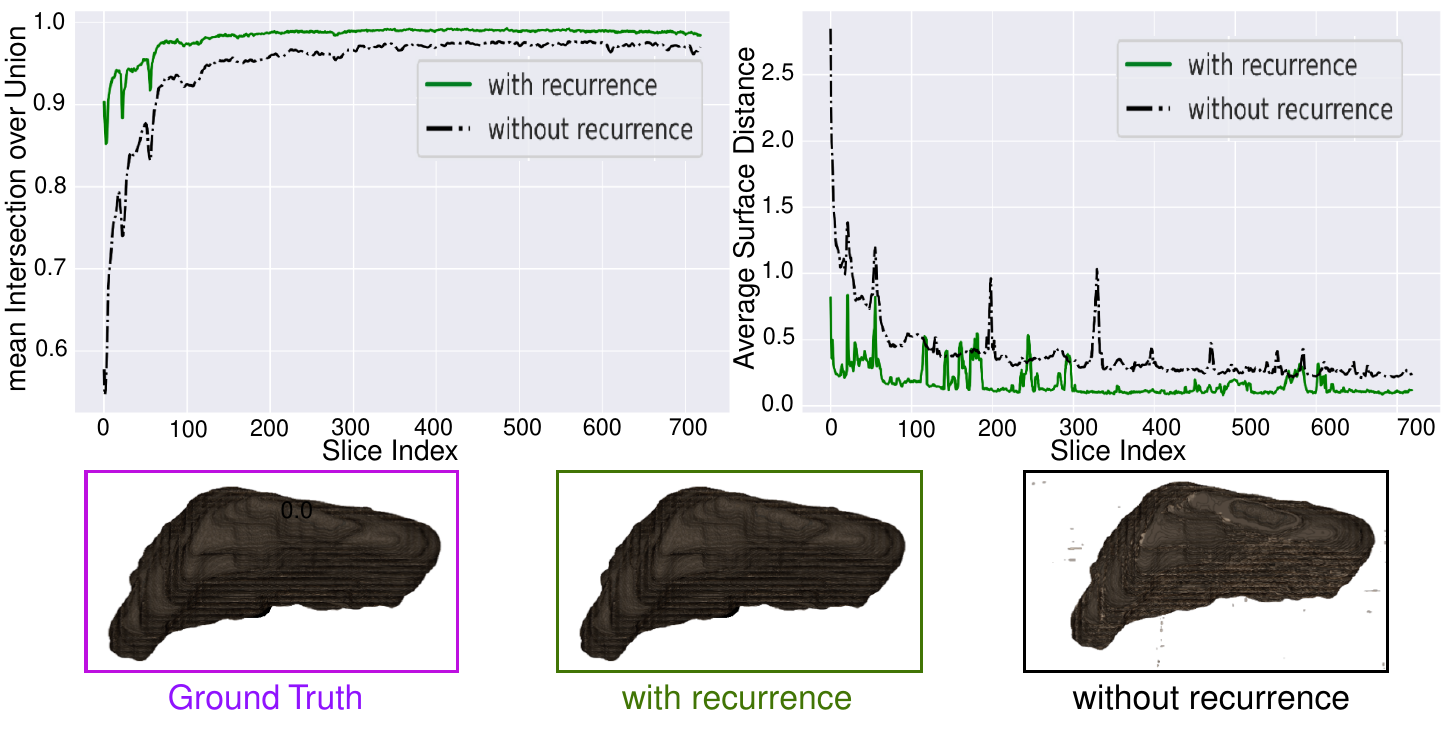}
    \caption{Ablation 2: network propagation performance with and without recurrence relation. A smoother and less noisy graph and volume prediction is obtained with recurrence relation module.}
    \label{RR}
\end{figure}

\textcolor{black}{\textit{Ablation 2: Recurrence Relationship}. We discarded the recurrence relationship, that is without feeding the previous} \textcolor{black}{output masks along with the current input sub-volume, but instead the network is relying only on the simulated expert intervention reference masks during the whole propagation process whenever they are provided. Herein, we show in Fig.~\ref{RR} that the network without the recurrence relationship produces a noisy and unsmooth boundaries.} \textcolor{black}{However, the network is still robust and consistent in its propagation over depth as reflected by the mIoU and ASD scores in Table~\ref{tab:RR}. We can conclude that the recurrence module provides smoothness to the propagation process because the network is updated constantly with new reference pseudo masks. Thereby, this module provides guidance to the network while the expert intervention minimal.}

\begin{table}[h!]
  \begin{center}
  \caption{Average ASD and mIoU scores and time to segment a volume.}
    \label{tab:RR}
    \begin{tabular}{c|c|c} 
      \textbf{\backslashbox{Metrics}{Method}} & \textbf{With Recurrence} & \textbf{Without Recurrence}\\
      \hline
      ASD score & \textbf{0.171} & 0.395  \\
      mIoU & \textbf{0.9827} & 0.948 \\
      Propagation Time (s) & 130 & \textbf{112} \\
    \end{tabular}
  \end{center}
\end{table}

\begin{figure}[ht]
 	\centering
 	\includegraphics[width=\linewidth]{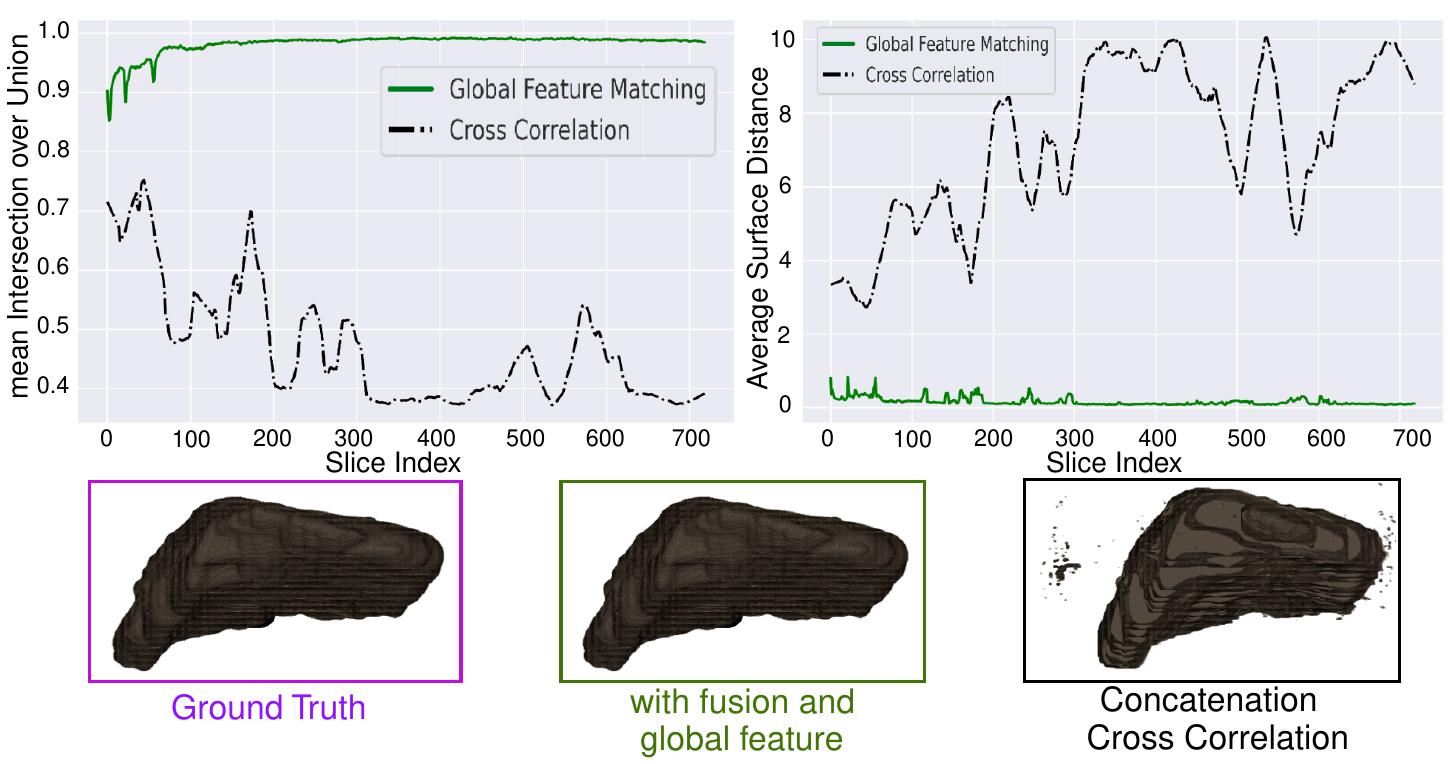}
      \caption{Ablation 3: network propagation performance with and without feature fusion an global feature matching.}
    \label{CC}
\end{figure}

\textcolor{black}{\textit{Ablation 3: Feature Fusion and Global Matching}. Next, we compared to a simplified method where we replaced the feature fusion (FF) with a simple concatenation (C) and we replaced the Global Feature Matching (GFM) with} \textcolor{black}{simple cross-correlation (CC) operator. The resultant scores reported in Table \ref{tab:CC}, show that the FF-GFM module plays a very important role in creating a common feature space between the images and the guiding masks. Fig.~\ref{CC} shows the localization of the features in image space and depth is noisy while using simple cross correlation for feature matching, in which feeding the BiCLSTM module with improper matched features, yields to a zero-order like interpolation output with unsmooth transition, also reflected in the in the ASD score.}

\begin{table}[h!]
  \begin{center}
    \caption{Average ASD and mIoU scores and time to segment a volume.}
    \label{tab:CC}
    \begin{tabular}{c|c|c} 
      \textbf{\backslashbox{Metrics}{Method}} & \textbf{FF-GFM} & \textbf{C-CC}\\
      \hline
      ASD score & \textbf{0.171} & 7.72  \\
      mIoU & \textbf{0.9827 }& 0.467 \\
      Propagation Time (s) & 130 & \textbf{122} \\
    \end{tabular}
  \end{center}
\end{table}

\begin{figure}[ht]
 	\centering
 	\includegraphics[width=0.95\linewidth]{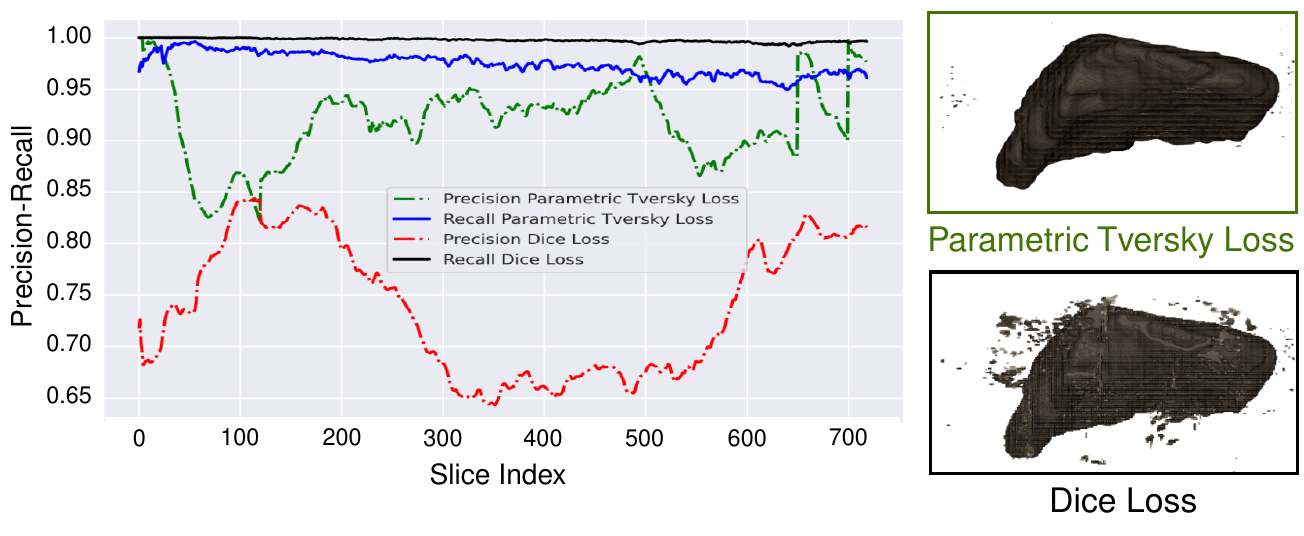}
    \caption{Ablation 4: loss function. Handling input and output voxel imbalance in muscle
segmentation. With Dice loss, high recall and low precision is recorded, while with Parametric Tversky loss, a trade-off between the recall and the precision is maintained.}
    \label{DC}
\end{figure}

\textcolor{black}{
%\textit{Ablation 4: Loss Function}. We trained our model with a Dice loss function that is equivalent to setting the weights $\alpha$ and $\beta$ of the equation \ref{objectiveLossFullSupervision} to $0.5$. On the other hand, we compare the results with our parametric Tversky-loss that figures out automatically the best weights $\alpha$ and $\beta$ to penalize the false positives and the false negatives pixels, respectively.}
\textit{Ablation 4: Loss Function}. We trained our model with 
our new parametric Tversky-loss that deduces automatically the best weights $\alpha$ and $\beta$ to penalize the false positives and the false negatives pixels in each sub-volume. We then compare against the same model with a Dice loss, which is equivalent to setting the weights $\alpha$ and $\beta$ of the equation \ref{objectiveLossFullSupervision} to $0.5$.} \textcolor{black}{Using dice loss, the trade-off between the precision and the recall is not properly established as shown in Table \ref{tab:DC} and Fig.~\ref{DC}. With Dice loss, the FN pixels are given more importance than the FP pixels due to the network bias and over-fitting towards background pixels as they are more numerous than the target muscle pixels. Handling this problem is very critical as muscle varies in size during its evolution in depth, that is we start with a small foreground muscle embedded in an abundance of background voxels. Therefore, our proposed parametric loss overcomes this issue by enforcing a desired trade-off between FNs and FPs as can be seen in Fig.~\ref{DC}, yielding better predictions for proper volume computation.}

% \textcolor{black}{Using dice loss, the trade-off of giving weights to pixels more than they deserve, false positives, and giving other} \textcolor{black}{pixels less than they deserve, false negatives, is not well accomplished, as all pixels are treated equally. Therefore, this yields to a high recall and low precision 

\begin{table}[h!]
  \begin{center}
      \caption{Precision-Recall Trade-off and time to segment a volume.}
    \label{tab:DC}
    \begin{tabular}{c|c|c} 
      \textbf{\backslashbox{Metrics}{Method}} & \textbf{Parametric Tversky Loss} & \textbf{Dice Loss}\\
      \hline
      Precision & \textbf{0.9187} & 0.738 \\
      Recall & \textbf{0.9744} & 0.997 \\
      Propagation Time (s) & 130 & \textbf{125} \\
    \end{tabular}
  \end{center}
\end{table}

\begin{figure}[ht]
 	\centering
 	\includegraphics[width=\linewidth]{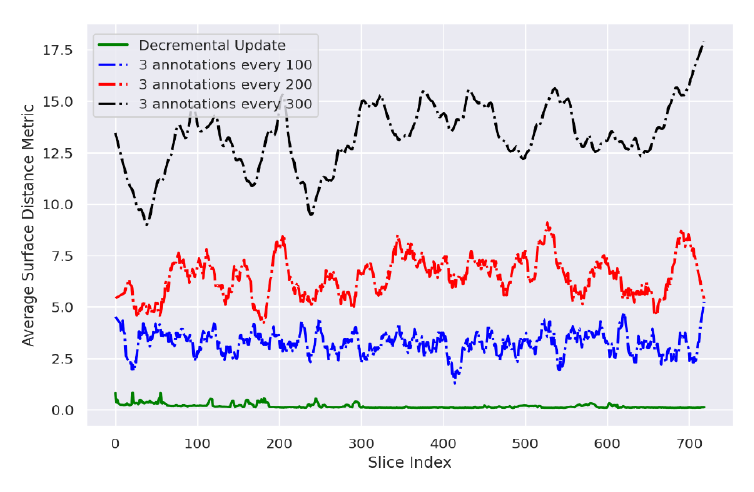}
    \caption{Ablation 5: supervision strategy: decremental vs fixed.}
    \label{DA}
\end{figure}

\textcolor{black}{\textit{Ablation 5: Supervision Mode}. Our model is trained in\emph{ decremental fashion }where the ground-truth is decrementally} 
\textcolor{black}{replaced by pseudo labels from one participant to the next while keeping a weak supervision of 3.5\% annotations over 29 training participants (a total of 1420 annotated masks). For example, the first participant will get 233 annotated slices} \textcolor{black}{which are uniformly and sparsely distributed over the full volume while second and the third will gets 116 and 58 annotations, and so on. Here, we show the effectiveness of decremental update by considering another three different settings: 1)} \textcolor{black}{we trained the model by providing supervision of 3 consecutive labels for each volume every 100 slices; 2) then, every 200 slices; 3) and finally,} \textcolor{black}{every 300 slices, each having a weak supervision percentage of 3.5\%, 2\% and 1\% respectively. \emph{With decremental update,} as our network is reinforced with more annotations at the first few epochs, it learn a better representation of the muscle being segmented as can be seen in Fig.~\ref{DA}, where we achieved a very good performance with $0.171~mm$ ASD score. We believe this is the result of less noisy pseudo segmentation maps that are reused in the next iterations. On the other hand, if we consider a fixed update under the same weak supervision annotation percentage (3.5\%), an average of $3.2~mm$ ASD score is obtained as shown in the blue graph. We also tested the performance once the percentage of annotation decrease further more to 2\% (red graph) and 1\% (black graph). We can see a more noisy pseudo-labeled are produced leading to unsmooth prediction with higher average ASD scores of $6.3~mm$ and $13.8~mm$.}

\begin{figure}[ht]
 	\centering
 	\includegraphics[width=0.75\linewidth]{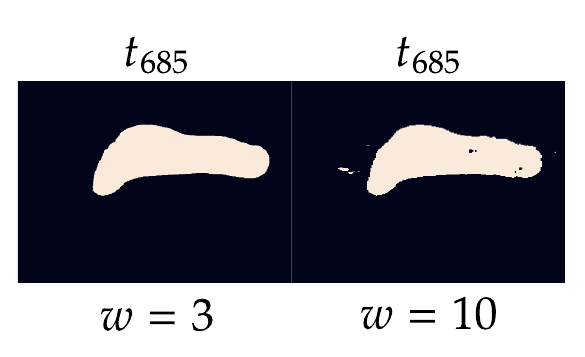}
    \caption{Ablation 6: Exploring kernel temporal depth. Here we can see a 2D slice at depth $t_{685}$ with $w=3$ and $w=10$ respectively.}
    \label{TWCh}
\end{figure}

\textcolor{black}{
\textit{Ablation 6: Exploring kernel temporal depth}. Here, we study the effect of varying the temporal depth window $w$. We compare two settings, one with $w=3$ and the other with $w=10$. Varying $w$ has a direct effect on the loss updates as we presented earlier in Fig.~\ref{supervisionWithFewShot}. For instance, with $w=10$, the sequential pseudo-labelling strategy become slightly noisier than with $w=3$, causing unnecessary errors and increasing the computational time as can be seen in Fig.~\ref{TWCh} and Table~\ref{tab:TWCh}.}

% With $w=10$, more parameters are expected to be learned and it requires far larger amounts of memory. t is prune to over-fitting if not well regularized. 

\begin{table}[h!]
  \begin{center}
     \caption{Average ASD and mIoU scores and time to segment a volume.}
    \label{tab:TWCh}
    \begin{tabular}{c|c|c} 
   \textbf{\backslashbox{Metrics}{Method}} & \textbf{$w=3$} & \textbf{$w=10$}\\
      \hline
      ASD score &\textbf{ 0.171} & 0.199  \\
      mIoU & \textbf{0.9827} & 0.9713 \\
      Propagation Time (s) & \textbf{130} & 134 \\
    \end{tabular}
  \end{center}
\end{table}

% Non-Learning Technique Studies.
\subsubsection{Non-learning Propagation Methods}
\begin{figure}[ht]
 	\centering
 	\includegraphics[width=0.80\linewidth]{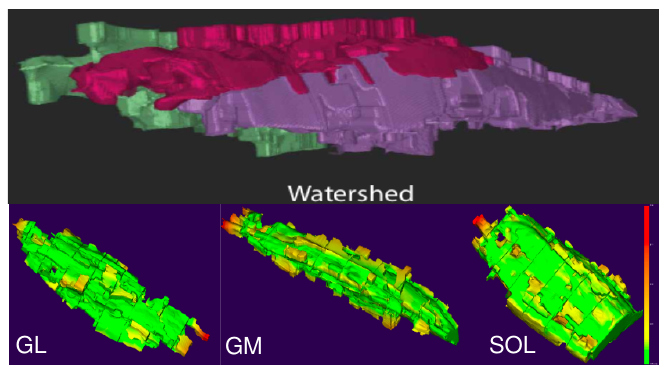}
    \caption{Top: predicted volume of GM (pink), GL (green) and SOL (violet) volume of WS method for one test patient. Bottom: the surface error maps.}
    \label{qualitativeAnalysisWS}
\end{figure}

%Learning methods to segment muscles in ultrasound images are not the only option. 
We perform a comparison with other popular non-learning interactive segmentation methods using 3.5\% range of binary masks as seeds. We follow approximately similar experimental protocol setting to IFSS-Net, however here each 100 slices, we provide 3 binary annotated masks as seeds.  
%capabilities, expected performance
FBS method does propagation over binary label masks only. It is an iterative morphological contour interpolator method that creates gradual change in the object. GFS and WS methods take into account the image content on top of the label-seeds. In addition to enforcing a smooth transition between the annotated slices, they push the segmented region's boundaries to coincide with the image contours. 
%why there are not popular
Most of our compared built-in approaches assume homogeneous areas of interest and well-defined image contours. Despite the competitive performance in other modalities, the assumptions above do not hold in the case of muscles in 3D US images, resulting in leakage. Without a specialized modification, the simple built-in implementation requires a large amount of labeled background and foreground seeds.
%Also highlight on the fact that they produce un-smooth temporal transition.
The application of FBS, GFS and WS methods on our dataset leads to the metrics reported in Table~\ref{tablesegmentation}. FBS gets a higher volumetric error than WS, because it depends on the closeness of the annotations to the edge. However, although WS has leakage, it has better volume estimation. In comparison to our IFSS-Net method, our network has learned to adapt to complex volume structure and to learn properly the shape properties over the entire volume, providing a better volume estimation while avoiding the weakness of FBS, GFS and WS. Qualitative results in Fig.~\ref{qualitativeAnalysis} and Fig.~\ref{qualitativeAnalysisWS}~evidence the daily challenges that experts are faced with when providing manual annotations for ultrasound images. The lack of well-defined edges and the little contrast makes it difficult to define the  muscle borders.

%  The lack of well-defined edges and the little contrast between regions of interest makes it difficult to define the segmentation mask borders. 

% \begin{table}[h!]
% \caption{Non-learning mask based propagation methods evaluation for muscle segmentation in 3D US scans.}
% \begin{tabular}{@{}cccccc@{}}
% \toprule
% Methods & Dice & mIoU & HDD & ASD & \% of volume error fraction \\ \midrule
% FBS & 0.918 & 0.849 & 16.6 & 1.057 & 10.78 \\
% GFS & 0.779 & 0.645 & 17.18 & 6.826 & 16.71 \\
% WS & 0.770 & 0.628 & 18.55 & 2.462 & 6.80 \\ \bottomrule
% \end{tabular}

% \label{tablesegmentation}
% \end{table}

\begin{table}[h!]
  \begin{center}
    \caption{Non-learning mask based propagation methods evaluation for muscle segmentation in 3D US scans.}
    \label{tablesegmentation}
    \resizebox{\columnwidth}{!}{%
    \begin{tabular}{c|c|c|c|c|c}
      \textbf{\backslashbox{Method}{Metrics}} & mIoU & Dice Coefficient & HDD  & ASD & VolErr\\
      \hline
        FBS & \textbf{0.918} & \textbf{0.849} & \textbf{16.6} &\textbf{1.057} &\underline{10.78} \\
        GFS & \underline{0.779} & \underline{0.645} & \underline{17.18} & 6.826 & 16.71 \\
        WS & 0.770 & 0.628 & 18.55 & \underline{2.462}  & \textbf{6.80} \\ 

    \end{tabular}}
  \end{center}
\end{table}

% \begin{figure}[ht]
%  	\centering
%  	\includegraphics[width=0.750\linewidth]{./Figures/PNoNLearning.pdf}
%     \caption{Top: predicted volume of GM (pink), GL (green) and SOL (violet) volume of WS method for one test patient. Bottom: the surface error maps.}
%     \label{qualitativeAnalysisWS}
% \end{figure}

% \section{Discussion}
% \textcolor{black}{new section: general highlight on the results}

\section{Conclusion}
In this paper, we proposed a new approach merging the benefits from expert interactions and deep-learning, dedicated to sequential or volumetric data \textcolor{black}{segmentation}. We deploy several strategies (Siamese network with subvolume recurrency, Bi-CLSTM, 3D ACS and pseudo-labelling) to exploit the spatiotemporal coherence of such data. The resultant IFFS-Net, allows propagating few-reference annotations over the entire volume/sequence while minimizing the expert efforts during training. We presented an in-depth evaluation of the muscle segmentation and volume estimation tasks ultrasound volumes.

% A shorter conclusion with perspectives
One of the perspectives of this work is the validation of our IFSS-Net over 3D freehand US volumes coming from children with Duchenne Muscular Dystrophy. With the disease progression, muscles become harder to segment as they are replaced by fatty tissues. Hence, some adaptions will be required.
%Hence, the expected outcomes of our network is not very high, as it has been totally trained over healthy participants and never seen any fatty tissues. 
One solution would be to fine-tune over a small set of DMD patients or train the network on second domain containing fatty tissues. Another solution would be to adapt the well known classification zero-shot learning paradigm for segmentation purposes. %The usability of  o
Our proposed methodology may also be useful for the segmentation of other anatomies requiring volume measurements and for other medical image analysis tasks dealing with sequential data. \textcolor{black}{Two final perspectives include testing the scalability of IFSS-Net across multiple anatomies %of interest
simultaneously and the generalization ability to multiple modalities.}

% \IEEEPARstart{T}{his} document is a template for \LaTeX.

%\subsection{Other Recommendations}

%\section*{Acknowledgment}
%\section{References}

% \input{Sections/complementaryMaterial}

\end{document}